\documentclass[10pt,twocolumn,letterpaper]{article}

\usepackage[pagenumbers]{cvpr} % To force page numbers, e.g. for an arXiv version

\definecolor{cvprblue}{rgb}{0.21,0.49,0.74}
\usepackage[pagebackref,breaklinks,colorlinks,allcolors=cvprblue]{hyperref}

\def\paperID{16148} % *** Enter the Paper ID here
\def\confName{CVPR}
\def\confYear{2025}
\usepackage{graphicx}
\usepackage{amsmath}
\usepackage{amssymb}
\usepackage{booktabs}
\usepackage{url}
\usepackage{float}

\usepackage[pagebackref,breaklinks,colorlinks]{hyperref}

\usepackage{pifont}
\usepackage{graphicx}
\usepackage{booktabs}
\usepackage{comment}
\usepackage{siunitx}
\usepackage{pdflscape}
\usepackage{soul}

\usepackage{multirow}
\newcommand{\mopicr}[1]{\textcolor{black}{#1}}
\newcommand{\mopi}[1]{\textcolor{black}{#1}}
\newcommand{\phflo}[1]{\textcolor{black}{#1}}

\newcommand{\revise}[1]{\textcolor{black}{#1}}
\newcommand{\gray}[1]{\textcolor{gray}{#1}}

 \newcommand{\myparagraph}[1]{\vspace{2pt}\noindent{\textbf{#1}}}
\newcommand{\xmark}{\ding{55}}
\usepackage{xcolor}
\usepackage{mathtools}
\usepackage{colortbl}
\usepackage[capitalize]{cleveref}
\crefname{section}{Sec.}{Secs.}
\Crefname{section}{Section}{Sections}
\Crefname{table}{Table}{Tables}
\crefname{table}{Tab.}{Tabs.}
\newcommand\blfootnote[1]{%
  \begingroup
  \renewcommand\thefootnote{}\footnote{#1}%
  \addtocounter{footnote}{-1}%
  \endgroup
}

\begin{document}

{
\title{

{\mopi{T-FAKE: Synthesizing Thermal Images for Facial Landmarking}}
\vspace{-0.3cm}
{
\author{
Philipp Flotho*\\
\small Saarland University\\
%{\tt\small Philipp.Flotho@uni-saarland.de}
\and
Moritz Piening*\\
\small Technical University of Berlin\\
%{\tt\small piening@math.tu-berlin.de}
\and
Anna Kukleva\\
\small MPI for Informatics, SIC
%{\tt\small piening@math.tu-berlin.de}
\and
Gabriele Steidl\\
\small Technical University of Berlin\\
%{\tt\small steidl@math.tu-berlin.de}
}
}}

\maketitle
%%%%%%%%% ABSTRACT
\begin{abstract}
Facial analysis is a key component in a wide range of applications such as healthcare, autonomous driving, and entertainment. Despite the availability of various facial RGB datasets, the thermal modality, which plays a crucial role in life sciences, medicine, and biometrics, has been largely overlooked. To address this gap, we introduce 
the T-FAKE dataset, a new large-scale synthetic thermal dataset with sparse and dense landmarks. To facilitate the creation of the dataset,  we propose a novel RGB2Thermal loss function, which enables the {domain-adaptive} transfer of RGB faces to thermal style. 
By utilizing the Wasserstein distance between thermal and RGB patches and the statistical analysis of clinical temperature distributions on faces, we ensure that the generated thermal images closely resemble real samples. 
Using RGB2Thermal style transfer based on our RGB2Thermal loss function, we create the {large-scale synthetic thermal} T-FAKE dataset$^\dagger$ \revise{with landmark and segmentation annotations.} 
Leveraging our novel T-FAKE dataset, probabilistic landmark prediction, and label adaptation networks, we demonstrate significant improvements in landmark detection methods on thermal images across different landmark conventions. Our models show excellent performance with both sparse 70-point landmarks and dense 478-point landmark annotations. \revise{Moreover, our RGB2Thermal loss leads to notable results in terms of perceptual evaluation and temperature prediction.}
%Our code and models {are available at} \url{https://github.com/phflot/tfake}. %Our code and models {are available at} \url{https://github.com/phflot/tfake}. 
\end{abstract}

\vspace{-.55cm}
%%%%%%%%% BODY TEXT
\section{Introduction}
\label{sec:intro}
\vspace{-0.2cm}

\begin{figure}
\centering
\includegraphics[width=0.9\linewidth]{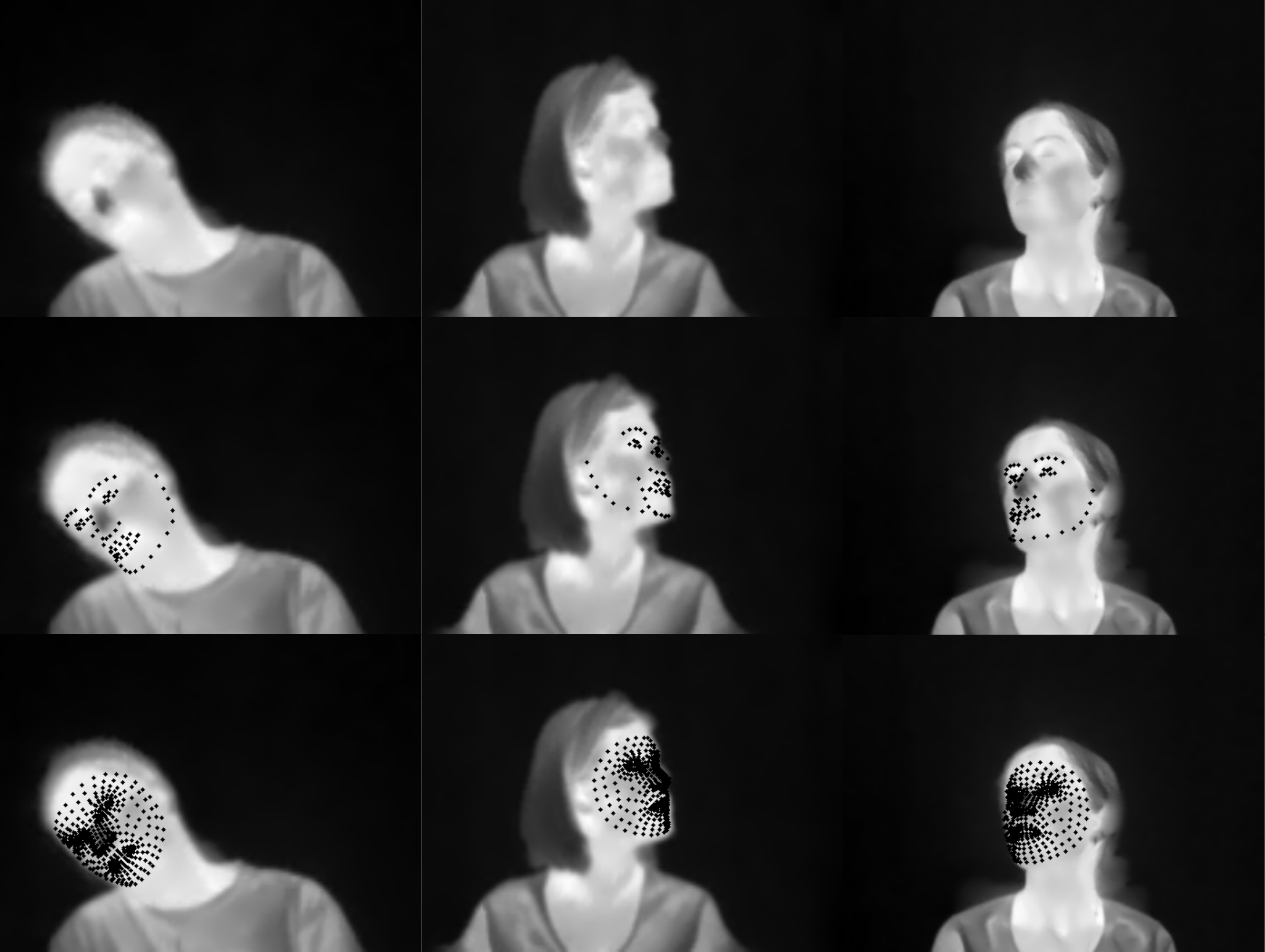}
\caption{Predictions of 70 (middle) and 478 (bottom) point facial landmarks on the challenging CHARLOTTE dataset which are estimated accurately by training a model with our synthetic data. }
\vspace{-0.6cm}
\label{fig:teaser}
\end{figure}

Functional thermal imaging is an acquisition modality with growing importance in many areas from life sciences over medicine to biometrics.
% \blfootnote{*Equal Contribution}\blfootnote{$^\dagger$\url{https://github.com/phflot/tfake}}
\blfootnote{*Equal Contribution; $^\dagger$\url{https://github.com/phflot/tfake}} Due to the close relation of body temperature with physiological states, thermal imaging of faces emerged as an important tool for screening infectious diseases \cite{quilty2020effectiveness, ng2005thermal, bitar2009international} and for the analysis of neurocognitive and affective states \cite{ioannou2014thermal}. As a consequence, there exists a growing need for thermal images with annotated facial landmarks.
However, in contrast to RGB images, annotated data for thermal
images of faces is still rarely available. 
For example, the
largest thermal datasets with facial landmark annotations contain only tens to
hundreds of individuals \cite{kopaczka2018thermal, flotho2021multimodal, poster2021large, ashrafi2022charlotte}, while the largest datasets for face detection for RGB images 
reached the mark of a million faces (of more than 500k individuals) 
with the MegaFace benchmark already in 2016 \cite{kemelmacher2016megaface}. 
Moreover, large synthetic 
RGB datasets, like FAKE \cite{wood2021fake}, appeared to be
 valuable 
for {sparse and dense} landmark detection training.
%\newline
Generally, the detection of facial landmarks is an important %{pre-processing} 
task for {further} facial analysis.
These 2D coordinates of corresponding points across faces are relevant for, e.g., face alignment \cite{wood20223d}, animation of 3D avatars \cite{lugaresi2019mediapipe}, 
motion magnification \cite{flotho2022lagrangian,FHSS2023} 
or deepfakes \cite{rana2022deepfake}. 
The most common convention for facial landmark annotations uses 68 points across the face and face boundary %, see, e.g., 
\cite{sagonas2016300}. 
Even though the 68-point convention remains challenging for thermal images \cite{poster2019examination, mallat2020facial}, 
%the importance of 
{there is additional demand for }dense landmarkers with more than 300 points. 
\cite{wood2021fake, wood20223d, lugaresi2019mediapipe}.

Faces in thermal images share structural similarities with RGB images. However, temporally varying environmental factors such as temperature and rain droplets, and physiological factors such as blood perfusion changes and vascular details lead to much higher variability in the appearance of faces \phflo{within individuals} \cite{ho2023ncku}. Sweating, flushing, or outdoor rain can result in time-dependent texture changes. Features that are very stable in RGB images such as the contrast of the nose and the eyes can become inverted when the nose or the face is cooled down.  
\revise{Thus, RGB landmarkers display decent results on thermal images \cite{flotho2021multimodal}, but show high failure rates and imprecise landmark predictions on challenging thermal images.}
Up to now, there exist only a few sparse thermal landmark detection methods \cite{kopaczka2018thermal, mallat2020facial, poster2021large, kuzdeuov2022sf, kuzdeuov2022tfw},
and we are unaware of dense thermal \mopi{or a multimodal RGB+Thermal} {landmarkers.}

In this paper, we aim to address this gap by introducing the T-FAKE dataset, the first large-scale synthetic thermal dataset featuring both sparse and dense landmarks for facial analysis. We propose to thermalize a synthetic RGB dataset using our novel RGB2Thermal loss function, which comprises three key components:
i) a supervised data term that controls the generation of thermal faces based on a small subset of RGB-thermal pairs,
ii) a Wasserstein distance term that aligns the patch distributions of the generated synthetic thermal images with those of real thermal images, and 
iii) a term incorporating prior information on clinical temperature statistics for different facial regions. {This loss enables our model to generalize beyond available lab-condition data to `in-the-wild' images. Figure \ref{fig:gan} shows T-FAKE samples and illustrates the advantage of our semi-supervised model over a supervised model trained solely with lab-condition images.}
To validate the quality of our T-FAKE dataset, we employ a probabilistic landmark prediction method \revise{with integrated face detection} by minimizing the negative log-likelihood function of the landmarks which are assumed to follow a Gaussian distribution as proposed in \cite{wood20223d}. Moreover, by introducing a learnable adapter between different landmark conventions, we unify various methods and datasets for evaluation. 

We thoroughly evaluate the landmark prediction method trained on our T-FAKE dataset on the challenging thermal dataset CHARLOTTE~\cite{ashrafi2022charlotte}, which includes controlled modulation
{of temperature, pose, and resolution} and on the established RGB dataset 
300W~\cite{sagonas2016300}. 
{
Our model demonstrates superior performance on the thermal landmark prediction benchmark compared to previous methods while maintaining performance on RGB images comparable to state-of-the-art RGB methods.}
\phflo{Figure}~\ref{fig:teaser} illustrates the performance of our model for sparse and dense landmarking. 

In summary, we provide the following contributions: 
\begin{itemize}
    \item %\ana
    {The first large-scale facial synthetic thermal dataset T-FAKE with \revise{70 sparse and 478 dense landmarks} \revise{and additional semantic segmentation masks};}
    \item %\ana
    {A novel RGB2Thermal loss to facilitate thermal image synthesis 
    {overcoming lab-recorded training data with limited image variability}}; 
    \item Training of the up-to-date first dense thermal landmarker and a state-of-the-art multimodal RGB+Thermal sparse landmarker 
    in combination with highly structured benchmarking across landmark conventions and modalities.
\end{itemize}

\begin{figure}
\includegraphics[width=.99\linewidth]{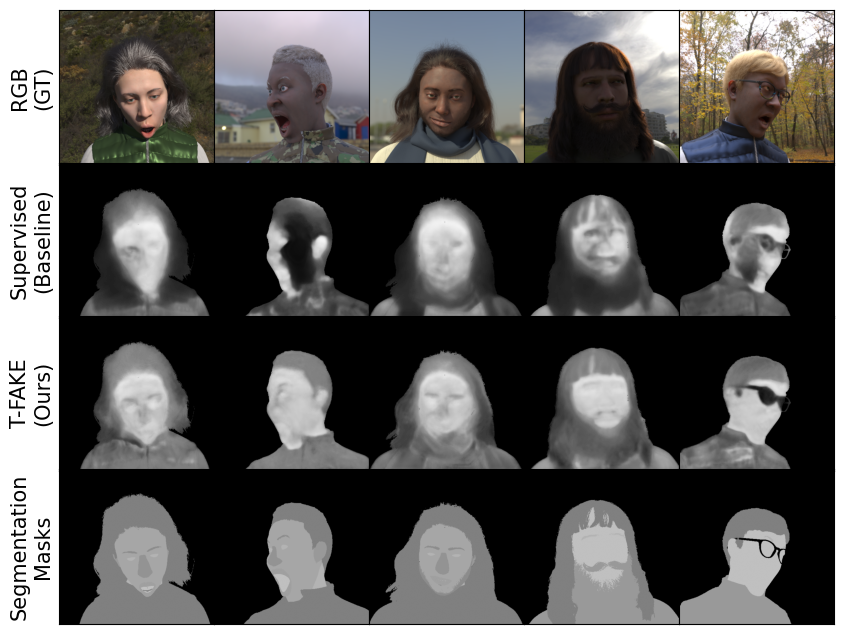}
\caption{{Challenging} frames from the FAKE dataset \cite{wood2021fake} before (first row) and after thermalization %with Pix2Pix \cite{isola2017image} 
without any regularization (second row) 
and our domain-adaptive model (third row). {Thermalized backgrounds are removed based on given segmentation.} The last row displays the segmentation masks with gray levels which are proportional to the reference temperature values used in regularizer \eqref{3}. {The first three images exemplify the `cold' setup with colder noses and the last two the `warm' setup with warmer noses. 
}}
\vspace{-0.5cm}
\label{fig:gan}
\end{figure}

\vspace{-0.2cm}

%------------------------------------------------------------------------
\section{Related Work}\label{sec:related}

\noindent\textbf{Facial RGB landmarking.}  
{Different annotation conventions exist for labeled landmarking datasets. An important distinction is between 2D annotations and 3DA-2D annotations which project 3D landmarks onto images \cite{deng2019menpo}. 2D is closer to human perception, but 3DA-2D enables annotation of occluded face regions.
}
The most widely used annotation convention defines 68-point {2D} landmarks {available for many facial alignment datasets} \cite{gross2010multi, koestinger2011annotated, zhu2017face, bulat2017far}. Sometimes points on the eye or other details are additionally added \cite{wood2021fake, ashrafi2022charlotte}. 
While an in-depth review of facial RGB landmarking is beyond the scope of this work, we introduce those that are relevant to thermal landmarking methods. Kazemi et al. \cite{kazemi2014one} introduced a {popular} traditional computer vision method for landmark prediction %that is still widely used 
via the dlib library \cite{dlib09}. %The model first localizes the face and then iteratively refines landmark positions with an ensemble of regression trees. 
Kowalski et al. \cite{kowalski2017deep} proposed the deep alignment network (DAN) which uses convolutional neural networks and iteratively refines {predictions}. Around the same time, Bulat et al. \cite{bulat2017far} proposed the facial alignment network (FA). Both methods predict heat maps of {landmark positions}. 
%from which coordinates are computed. 
Explicit integration of uncertainty quantification into a %deep 
{landmark prediction model} 
via the LUVLi loss has been proposed by Kumar et al. \cite{kumar2020luvli}.
{Recently, image annotation quality} became {a limiting factor }
%of modern algorithms 
which led to the emergence of synthetic data with high-quality ground truth \cite{melzi2024frcsyn}. 
Wood et al. \cite{wood2021fake, wood20223d} trained {3DA-2D} landmarkers { with excellent performance using their proposed synthetic} 
FAKE dataset with 100k frames of {at least 2k individuals}. \revise{Note that most landmarkers are trained on facial bounding boxes. As a result, they are either evaluated in combination with a face detector or on pre-defined ground truth bounding boxes. Here, we focus on the combined evaluation of face and landmark detection, a harder, but more practical task.}
{\scriptsize{
\begin{table*}
\centering
\resizebox{\textwidth}{!}{
\begin{tabular}{llrrrrrrr}
Dataset Name & Year & DS Available & Model Available & Landmarks & Size (F/S) & Distance (\SI{}{\cm}) &\revise{Out-of-Lab}\\
\hline\hline
Aachen \cite{kopaczka2018thermal} & 2018 & \checkmark & \checkmark & 68 & 2935/90 & 90& \xmark\\
ARL-MMFD \cite{zhang2019synthesis} & 2019 & (request) & \xmark & {6} & {888}/111 & 250, 500, 750& \xmark \\
DRIVE-IN \cite{flotho2021multimodal} & 2021 & \xmark & \xmark & 68 and {478} & {$\sim$7-8M}/436 & $\sim$80-150 & \checkmark\\
ARL-VTF \cite{poster2021large} & 2021 & \xmark$^*$ & \xmark & {6} & {$500$k}/395 & 210 & \xmark\\
TFW \cite{kuzdeuov2022tfw} & 2022 & \checkmark & \checkmark & 5 & 9982/147 & $\geq$100 & \checkmark\\
SF-TL54 \cite{kuzdeuov2022sf} & 2022 & \checkmark & \checkmark & 54 & 2556/142 & 100 & \xmark\\
CHARLOTTE \cite{ashrafi2022charlotte} & 2022 & \checkmark & \xmark & {72} (43 profile) & {7684/10} & 100-660 & \xmark\\
%NCKU-VTF \cite{ho2023ncku} & 2023 & (request) & \xmark & -- & 6000/ & ---- \\
T-FAKE (ours) & 2024 & \checkmark & \checkmark & 70 and {478} &  200k/{$\geq$}2k & $\sim$80-150 & \checkmark
\end{tabular}
}
\label{tab:results}
\caption{Thermal landmark dataset comparison. \revise{Indicated are the number of frames (F) and subjects (S), camera distance, and if recorded outside of a controlled laboratory. Unavailable data are marked by \xmark$ $ and \xmark$^*$ indicates no response received from corresponding authors.} % data on request without response to our requests}. %\textcolor{red}{Note that for CHARLOTTE we report the total number of available 16bit image - annotation pairs in the final dataset rather than the reported number of acquired frames.} 
}
\vspace{-0.5cm}
\label{tab:ds}
\end{table*}
}}

\noindent\textbf{Thermal landmarking.} 
Many tailored approaches for thermal landmarking use similar architectures as RGB landmarkers and are mainly developed alongside the publication of new datasets, see  Table \ref{tab:ds}. 
Kuzdeuov et al. \cite{kuzdeuov2022tfw} released the TFW dataset with 
%{\st{a total number of 9982 images from 147 subjects annotated}}
with 5 facial landmarks and bounding boxes and trained YOLOv5 networks for landmark prediction. The SF-TL54 dataset \cite{kuzdeuov2022sf} %{\st{consists of 54 {{2D}} landmarks and 2556 frames of 142 subjects and}} 
has been used for {2D} landmark prediction with a dlib and a U-net model. Kopaczka et al. \cite{kopaczka2018thermal} use a DAN for their AACHEN dataset with %{\st{annotated 2935 images of 90 subjects and}} 
68-point {{2D}} landmarks. % and emotion labels.
A highly structured {2D landmarking} dataset is the CHARLOTTE thermal face dataset {with controlled temperature changes} 
%{\st{with 10 subjects and 10k images}} 
by Ashrafi et al \cite{ashrafi2022charlotte}.
%{\st{, who annotated their dataset with 72-point {{2D}} landmarks}. 
It contains around 10k images in total and 7684 images with annotations. %While the dataset consists of only 10 individuals, they methodologically record various difficult head-poses as well as distances to the camera and image resolutions in 10k images.  
In contrast to these publicly available datasets, datasets with limited or no public availability include the 
the ARL-VTF dataset \cite{poster2021large} 
and a dataset recorded in SARS-CoV-2 drive-in stations (DRIVE-IN) \cite{flotho2021multimodal} (six minutes recordings at $50$Hz of 436 subjects). The authors of the DRIVE-IN dataset found that a stack of pre-processing functions already leads to a good performance of existing RGB landmark models on thermal images.
\revise{Note that such thermal datasets are often captured with uncooled microbolometers with automatic radiometric calibration. This leads to comparable temperature values but may introduce small measurement errors.}
Beyond thermal images, Poster et al. \cite{poster2021visible} proposed to use transfer learning on a siamese DAN to improve thermal landmark prediction with RGB facial landmarks. Mallat et al. \cite{mallat2020facial} used RGB2Thermal image translation (`thermalization'), active appearance models, and deep alignment networks of RGB datasets for thermal landmark prediction. {In contrast to the previous work, our dataset contains thermal images of {at least} 2k individuals that are annotated with sparse 70 {3DA-2D} and dense 478 {2D} point landmarks.}

\noindent\textbf{Thermalization.}  In addition to thermal landmark datasets, various datasets of paired and aligned thermal and RGB images without annotations exist. These can be employed for the training of thermalization models. {In contrast to the rich literature on Thermal2RGB synthesis \cite{riggan2018thermal, zhang2018tv, nair2023t2v}, the literature on RGB2Thermal image synthesis using such datasets is more limited.} 
Mallat et al. thermalized facial images by minimizing a perceptual loss \cite{mallat2020facial} using the VIS-TH dataset \cite{mallat2018benchmark}.
 Similarly, generative adversarial networks (GANs) have been used for {Thermal2RGB synthesis} \cite{zhang2019synthesis, ho2023ncku} using the NCKU-VTF \cite{ho2023ncku} dataset and the ARL-MMDF \cite{zhang2019synthesis} dataset containing facial image data. In contrast to the limited public availability of these two datasets, a publicly available dataset with paired facial images is the multi-modal disguise \textit{Sejong face database} (SEJONG) with 100 subjects, 15 disguise variations and 4849 RGB+Thermal image pairs by Cheema and Moon \cite{cheema2021sejong}.
 {Note that all these datasets and resulting models are mostly restricted to frontal views of faces with controlled lighting.}
 Thermalization of general scenes has been achieved by Kniaz et al. \cite{kniaz2018thermalgan} with a modified Pix2Pix GAN based on the 5098 {triplets of RGB images, thermal images, and segmentation masks} in the %publicly available 
 ThermalWorld dataset.
%pairs displaying various object classes.
{In contrast to previous works, we employ a semi-supervised RGB2Thermal loss using a semantic segmentation to improve generalization {beyond lab-condition frontal views to `in-the-wild' images}.}
\begin{figure}
\centering
\includegraphics[width=.99\linewidth]{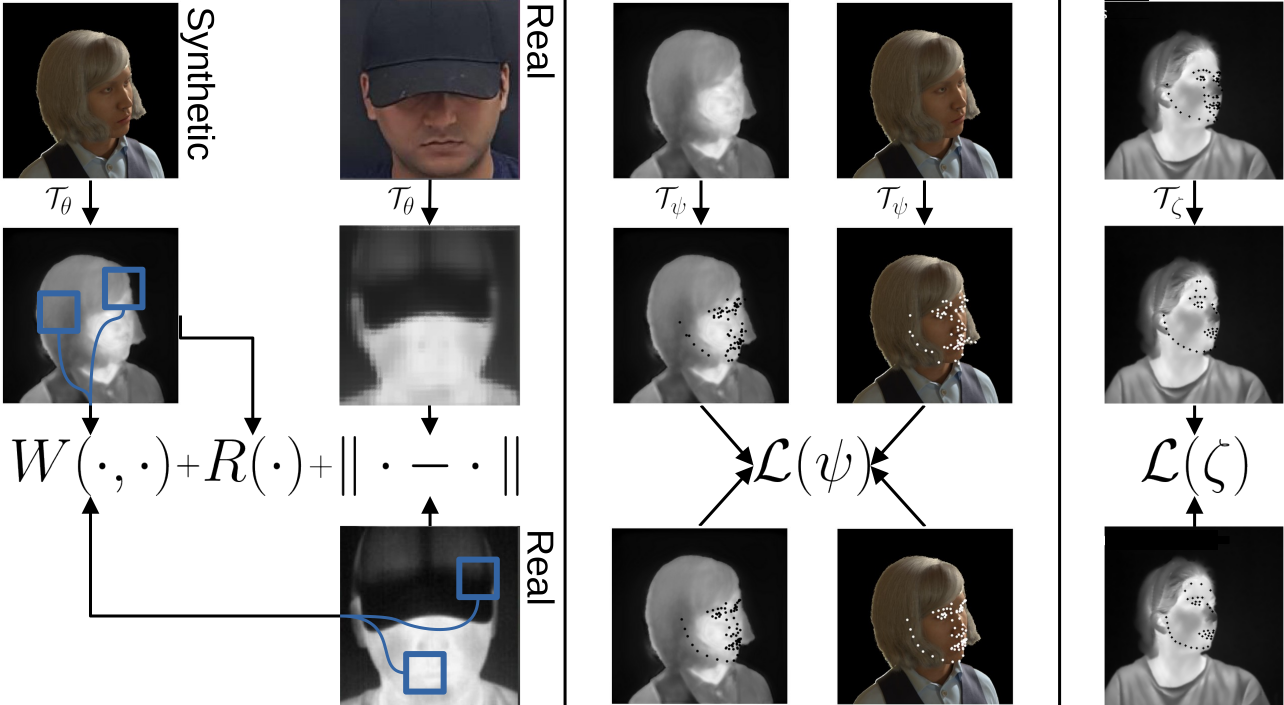}
\caption{Three steps of our training pipeline (models and their loss functions):  1. thermalization network $\mathcal{T}_\theta$ with loss \eqref{1} - \eqref{3}.  
2. landmark prediction network $\mathcal{T}_\psi$
with \eqref{loss_psi}. 3. landmark adaptation network $\mathcal{T}_\zeta$ with \eqref{loss_zeta}. We show black dots for thermal landmarks and white dots for RGB landmarks. Note the landmark adaptation on the forehead in the last step.}
\vspace{-.5cm}
\label{fig:training_pipeline}
\end{figure}

\section{T-FAKE Dataset}
%\anna{be aware of misalignment}
Recently, the success of facial landmarkers trained solely on the FAKE dataset \cite{wood2021fake}, a synthetic facial landmarking 
dataset, has demonstrated the potential of synthetic data for the task.
% for landmarking.
We aim to transfer this success story to the largely overlooked thermal domain and the multimodal setting by introducing our synthetic thermal landmarking dataset T-FAKE featuring {at least} 2k unique subjects. 
%\newline
In particular, we train a semi-supervised image translation model for the thermalization of RGB images using real and synthetic data. To account for {data variations} %in the data 
in terms of poses, {shadows,} and expressions, we employ a semi-supervised loss.

%\vspace{-.1cm}
\subsection{Thermalization}
%\vspace{-0.6cm}

%\begin{enumerate}
%    \item \textbf{Thermalization} of the FAKE dataset to obtain the landmarked thermal synthetic dataset T-FAKE
    %by employing an image translation model whose semi-supervised loss accounts for
     %variations of the data in terms of poses, hairstyle, etc.
%    \vspace{-0.2cm}
    %\item \textbf{Landmark detection} in thermal images
    %by learning a model  
    % based on both the FAKE and T-FAKE datasets.
     %\vspace{-0.3cm}
    %\item \textbf{Label adaptation} to
    %by training a NN that can 
    %translate between incompatible landmark conventions.
% \end{enumerate}
%For an overview of the three networks and their loss functions, in particular, the required inputs for the losses, see Fig. \ref{fig:training_pipeline}.
%\anna{Again repeat one sentence motivation why we need thermalization: We introduce our T-FAKE dataset and we propose thermalization of synthetic RGB dataset. It's alright to repeat the same motivation over and over again in different words}
To generate our large-scale synthetic T-FAKE dataset,
% we need a thermalization model. 
we {propose} a thermalization {framework} (\phflo{see Figure \ref{fig:training_pipeline})}. 
{First,} we train our network in a semi-supervised manner 
based on training pairs of RGB+Thermal images 
$
(\mathbf X_{\text{ RGB}}, \mathbf X_{\text{T}}) 
= (X_{\text{ RGB}}^n,X_{\text{T}}^n)_{n=1}^N 
$
from the lab-recorded SEJONG % \textcolor{red}{and  DRIVE-IN} 
dataset. 
Further, we use the synthetic RGB images from FAKE 
$
\mathbf{X}_{\text{FAKE}} = (X_{\text{FAKE}}^m)_{m=1}^M
$
for which no thermal counterpart exits.
% Indeed
The thermalization of these data 
appears to be difficult due to the large differences between real-world lab-recorded and synthetic in-the-wild data when facing thermalization tasks. % due to, e.g., shadows, poses, or expressions. 
% We propose to construct an U-net \cite{ronneberger2015unet} $T_\theta$ for RGB2Thermal translation by minimizing the loss function
For RGB2Thermal translation, we {optimize} a U-net \cite{ronneberger2015unet} $T_\theta$ by minimizing the loss function
\begin{align}
    \mathcal{L}(\theta) 
    &\coloneqq
    \frac1N\sum_{n=1}^N {\| T_\theta(X_{\text{ RGB}}^n)- X_\text{T}^n\|}_2^2
    \label{1}\\
    &% + \operatorname{W}(T_\theta(\mathbf X_{\text{ RGB}}),\mathbf X_{\text{T}}) 
     + \lambda_W \operatorname{W}(T_\theta (\mathbf X_{\text{FAKE}}),\mathbf X_{\text{T}})
    \label{2} \\
   & + \lambda_R \frac1M \sum_{m=1}^M \operatorname{R}(T_\theta (X_{\text{FAKE}}^m), \quad 
   \lambda_W,\lambda_R >0 \label{3}.   
\end{align}
This loss allows us to train a supervised image translation model from paired data while ensuring domain-generalization of the model using unpaired data. The first summand \eqref{1} penalizes the deviation of the generated thermal image from the true one and serves as a reconstruction loss for known image pairs. %\anna{Expand a bit on the first term and add motivation why to have this loss. }
The second summand  \eqref{2} exploits %\anna{summands? sure that not summand? } 
the Wasserstein distance 
on the patch distribution
of thermalized FAKE images and natural thermal images. 
This  style loss ensures that a generated image
displays the thermal image statistics.
To be more precise,
recall that the squared Wasserstein-2 distance $\mathcal W_2 ^2$ between empirical measures $\mu = \frac1K\sum_{k=1}^K \delta_{x_k}$
and 
$\nu = \frac1L \sum_{l=1}^L \delta_{y_l}$ {for Dirac probability measures $\delta_{\bullet}$} is given by
\begin{align}
\mathcal W_2 ^2 (\mu,\nu)\hspace{-.1cm} \coloneqq \hspace{-.1cm} \inf_{\pi \in \Pi} F(\pi) \hspace{-.1cm} \coloneqq \hspace{-.1cm} \inf_{\pi \in \Pi} \sum_{k=1}^K \sum_{l=1}^L \|x_k - y_l\|^2 \pi_{k, l},
\end{align}
where 
$\Pi$ denotes the set of non-negative $K\times L$ matrices 
which rows and columns sum up to $1/K$ and $1/L$, respectively. Calculation with this distance requires solving costly optimization problems. As a remedy, the entropy-regularized Wasserstein distance
\begin{align}
\mathcal W_{2, E} ^2 (\mu,\nu) \hspace{-2pt}
&\coloneqq \hspace{-2pt}\inf_{\pi \in \Pi} \big\{ 
F(\pi)\hspace{-2pt}+\hspace{-2pt}\lambda_E \hspace{-2pt}
\sum_{k=1}^K \hspace{-2pt}\sum_{l=1}^L \hspace{-2pt}\pi_{k, l} \hspace{-1pt}
%{\sum_{k, l=1}^{K, L}} %\sum_{l=1}^L \pi_{k, l} 
\log \left( \pi_{k, l}\right) \big\}
\end{align}
with some (small) regularization $\lambda_E> 0$ has been proposed \cite{cuturi2013sinkhorn}. {This term} and {its} gradients can efficiently be calculated via the Sinkhorn algorithm \cite{feydy2019interpolating, piening2023learning}.
%\ank{it feels that some connection is missing between previous and the next sentence}
{To use this loss for thermal translation, we focus on the (subsampled) patch distributions of the generated and real images. For that purpose, we now} extract (overlapping) 
 image patches {by defining the patch extractor} $P_{i}~X~\coloneqq ~X[i_1:~i_1+s,~i_2~:~i_2~+~s]$, 
 $i =(i_1,i_2) \in I$ of size $s \times s$ and
 {considering} the  empirical measures generated by the patch distributions
 of  $T_\theta(\mathbf X_{\text{FAKE}})$,
\begin{equation}
\mu_{\text{FAKE}} \coloneqq \tfrac{1}{{M}} \tfrac{1}{|I|} \sum_{{m}=1}^{{M}} \sum_{i=1}^{|I|} \delta_{P_{i} \left( T_\theta 
(X_{\text{FAKE}}^n) \right)},
\end{equation}
%and similarly  $\mu_{\text{F-RGB}}$ from $T_\theta(\mathbf X_{\text{F-RGB}})$
and similarly  $\mu_{\text{T}}$  from $X_{\text{T}}$.
Now we penalize in \eqref{2}  the term
\begin{align}
%\operatorname{W}(T_\theta(\mathbf X_{\text{ RGB}}),\mathbf X_{\text{T}}) 
%&\coloneqq
%\mathcal W_{2, E}^2(\mu_{\text{RGB}},\mu_{\text{T}}),
%\\
\operatorname{W}(T_\theta(\mathbf X_{\text{FAKE}}),\mathbf X_{\text{T}}) 
&\coloneqq
\mathcal W_{2, E}^2(\mu_{\text{FAKE}},\mu_{\text{T}}).
\end{align}
{During training}, 
%\ank{soory, but what is the numerical examples? experiments? do we have ablations on that?}  
we apply a \emph{multiscale approach} by applying this regularizer over multiple image scales and summing {over} the results.
As a result, the patch statistics 
of downscaled images are incorporated into the empirical loss. 
{The proposed patch distribution matching-based regularization was inspired by its application in related tasks such as}
%Bayesian inverse problems
image reconstruction \cite{altekruger2023patchnr, piening2023learning}, 
style transfer \cite{leclaire2021stochastic},
texture reconstruction \cite{hertrich2022wasserstein, altekruger2023wppnets}, and image translation \cite{isola2017image}.
%\ank{the last sentence is too much. it rather tells - okey, we actually do quite incremental stuff here..  reformulating into sth closer to "our patch distribution matching was inspired by ..."}
%\anna{better to expand on how exactly you do multiscale approach. + below it is rather some related work subsection. It should be shortened to one sentence max.}

%Penalizers of Wasserstein distances on patch distributions
%were used for texture generation and style transfer in \cite{leclaire2021stochastic} and
%for the construction of regularizing terms in Bayesian inverse problems in
%\cite{altekruger2023wppnets,hertrich2022wasserstein}. 
%Alternatively, regularizers based on normalizing flows on patch distributions were applied for image reconstruction in \cite{altekruger2023patchnr}, and patch discriminators based on GANs were proposed for image translation in \cite{isola2017image}.
 %\textcolor{red}{Our prior also resembles the patch-based Pix2Pix discriminator \cite{isola2017image}, 
 %but allows for more stable training by avoiding adversarial training.}

%{\color{red}}  More precisely, we deal with the entropically regularized Wasserstein distances \cite{}.
% Sinkhhorn algorithm \cite{feydy2019interpolating} 
%} 

\mopi{Finally, the {distribution of temperature} across different regions of the face and body are well-studied \cite{sonkusare2021data, cosic2022temperatures}, e.g., warmer areas around the eyes, colder scalp hair and glasses blocking thermal heat. Thus, the third summand \eqref{3} incorporates such knowledge about the expected temperature in different facial areas.} To be specific, we make use of a given semantic segmentation of the synthetic dataset of the image $\mathbf X_{\text{F-RGB}}$ into pixels corresponding to 18 regions $S_i$, e.g., `background', `nose', `headwear' and so on. %\anna{missing motivation, to abrupt move to the technical details}
%("background", "skin", "nose", "eyes", "brows", "ears", "lips, "mouth interior", "hair", "beard", "clothing", "glasses", "headwear" and "facewear"). 
Moreover, let $T_i$ be the average temperature of the hardcoded reference temperatures in the region $S_i$ based on measurements presented in empirical studies \cite{cosic2022temperatures, sonkusare2021data, ashrafi2022charlotte}.
Then, we compute
\begin{equation} 
    \operatorname{R} \left(T_\theta(\mathbf X_{\text{FAKE}}) \right) 
    =   
    \sum_{m=1}^M \sum_{i=1}^{18} \omega_i (\overline{S_i(T_\theta(X_{\text{FAKE}}^m)}) - T_i)^2, 
\end{equation}
where the weights $\omega_i$ are proportional to the region sizes and $\overline{S_i}$ is the mean temperature in the $i$-th region. 
Note that facial temperatures depend on the environmental conditions. Hence, we train $T_\theta$ two times with two different reference temperature sets to simulate `cold' and `warm' environmental conditions.
% changed order of the figures, here:
\phflo{The segmentations are shown in Figure}~\ref{fig:gan} (bottom) and different regularizers in \phflo{Figure}~\ref{fig:training_pipeline} (left). %are shown

 Predictions of the normalized temperature with our model compared to a model without regularization are shown in \phflo{Figure}~\ref{fig:gan}.  
 The advantages of our regularization in a domain-adapted setting are clearly visible. The use of regularization reduces image artifacts. The patch regularizer \eqref{2} promotes natural texture synthesis and because of our segmentation regularizer \eqref{3} temperature differences between facial areas are more pronounced. Hence, our method allows for the adaptation of 
 images with shadows, various skin colors, and distinctive facial expressions. In particular, such features may be underrepresented in facial thermal imaging datasets recorded in lab conditions. 
% \anna{this section ends with thermalizing the dataset} \mopi{What do you mean?}
 Note that the resulting domain adaptation needs for synthetic data 
 %\mopi{and the access to ground truth segmentations} 
 {solved by a semi-supervised approach based on the access to ground truth segmentations} differentiates our task from the 
 {RGB2Thermal tasks in \cite{kniaz2018thermalgan, mallat2020facial} 
 and (supervised) Thermal2RGB tasks in \cite{zhang2018tv, nair2023t2v, ho2023ncku}. Generalization limitations of supervised models are \phflo{discussed in the supplementary}.}}
 %thermalization tasks in \cite{kniaz2018thermalgan, zhang2019synthesis, mallat2020facial, ho2023ncku}.} %in related works
\vspace{-0.1cm}
\subsection{The Dataset}

 Having trained  $T_\theta$, we use it to thermalize the FAKE
dataset which results in the thermal dataset T-FAKE. {For each thermal condition, the} dataset contains 100k images featuring {at least }2k unique subjects. Each image is available in a simulated `cold' and a `warm' environment {adding up to 200k images}.  This number of frames and subjects surpasses the numbers in other thermal landmarking datasets \cite{kopaczka2018thermal, zhang2019synthesis, flotho2021multimodal, poster2021large, kuzdeuov2022tfw,kuzdeuov2022sf, ashrafi2022charlotte} known to the authors. In contrast to thermal landmarking datasets with a fixed distance between the subjects and the camera \cite{kopaczka2018thermal, poster2021large, kuzdeuov2022sf}, the distance varies within {an estimated} range of 80cm to 150cm for the T-FAKE dataset. Moreover, the synthesized images are based on facial scans with a wide range of age and ethnicity and a balanced number of people identifying as male or female \cite{wood2021fake}. In addition to the original ground truth sparse {3DA-2D} landmark annotations, we add dense {2D} annotations by transferring Mediapipe \cite{lugaresi2019mediapipe} predictions for the FAKE to the T-FAKE dataset. This makes our dataset the first densely annotated thermal landmarking dataset. {While we found many %of the 
large thermal landmarking datasets to be non-public \cite{zhang2019synthesis, flotho2021multimodal, poster2021large}, our complete dataset with both thermal conditions and annotations is available for download. 
%{Contrary to other nonpublic thermal landmarking datasets \mbox{\cite{zhang2019synthesis, flotho2021multimodal, poster2021large}}, we plan to make our dataset available for download upon acceptance.}
} 
%\ank{this sentence should be revised before arxiv - sth like "our dataset is available for download.." }
%\textcolor{red}{Related to multimodal comment:} 
{In combination with the original FAKE dataset, we can train multimodal facial landmarkers jointly on RGB and thermal images as described in the next section.
%\st{As a result of all these features, the T-FAKE dataset is an attractive dataset for learning a landmark prediction model as described in the next section.}
}  
Table \ref{tab:ds} displays {a comparison} between thermal landmarking datasets.  
%detailed
%\begin{figure}
%\includegraphics[width=1\linewidth]{Thermal/images/Seg-Viz.png}
%\caption{Synthetic RGB images from the FAKE dataset %\cite{wood2021fake} (top) and their segmentation masks (bottom), where the gray level of each segment corresponds to the reference temperature values used in \eqref{loss_psi}.}
%\label{fig:seg}
%\end{figure}

\section{Facial Landmarking}
%---------------------------------------------
Learning with synthetic images has proven to be very useful in facial landmarking \cite{zhang2019synthesis, wood2021fake, wood20223d}. Based on {the proposed} T-FAKE and the original FAKE dataset, we {can train} a supervised thermal or multimodal landmark prediction model for thermal and RGB images using synthetic data only. Moreover, we {adapt a} model to new annotation conventions by training a small-scale label adaptation model. The data flow for {these models} is visualized in \phflo{Figure}~\ref{fig:training_pipeline}.
\vspace{-0.1cm}
\subsection{Landmark Prediction}
We apply the construction in \cite{wood20223d} and learn a landmark prediction model through probabilistic landmark regression from given paired data
$\big(X^j, \mathbf{y}_j \big)_{j=1}^J$,
where $X^j \in \text{FAKE} \cup \text{T-FAKE}$ and
$\mathbf{y}_j \coloneqq (y^j_l)_{l=1}^L$, $y^j_l \in \mathbb R^2$ are the corresponding $L$ landmarks.
We learn a network $T_\psi$ mapping from the space of images  to the space of $L$-fold
two-dimensional Gaussian distributions  $\big( \mathcal N(\mu_l, \sigma_l I_2) \big)_{l=1}^L$.
Here we employ a MobileNet V2 architecture for landmark prediction \cite{sandler2018mobilenetv2}. 
Recalling that the Gaussian distribution 
$\mathcal N(\mu, \sigma I_2)$ 
has the \emph{negative log-likelihood function}
\begin{equation}
-  \log p (y|\mu,\sigma) =   \log \left(2 \pi \sigma^2\right) + \frac{\|\mu - y\|_2^2}{2 \sigma^2},
\end{equation}
the authors of \cite{wood20223d} proposed to learn the
network $T_\psi \coloneqq (T_\psi^1,T_\psi^2)$ with $X \mapsto (\boldsymbol{\mu}, \boldsymbol{\sigma}^2)$, where
$\boldsymbol{\mu} = (\mu_l)_{l=1}^L \in (\mathbb R^2)^L$
and
$\boldsymbol{\sigma}^2 = (\sigma_l^2)_{l=1}^L \in \mathbb R^L$
by minimizing the loss function
{
\small{
\begin{align}
	\mathcal{L}(\psi) &\coloneqq 
 \sum_{l=1}^L %w_l 
 \Big( 
 \sum_{j=1}^J 
 \log T^2_{\psi,l} (X^j)
  + \frac{\|T^1_{\psi,l}(X^j) - y_l^j\|_2^2}{2 \, T^2_{\psi,l} (X^j)
 } \Big).
 \label{loss_psi}
\end{align}
}
}
 %To additionally balance different face regions, we included fixed weights $w_l$ as a hyperparameter for each landmark location.  
During training, we clip $T^2_{\psi,l} \geq \varepsilon >0$ to avoid numerical instability due to too small variances.
The first sum controls the location accuracy and the second one the uncertainty prediction.  Having learned the network $T_\psi$, it produces not only the
expected values $\boldsymbol \mu$ of the landmarks, but 
also a measure for its uncertainty $\boldsymbol \sigma^2$. \revise{We threshold the average uncertainty score of multiple sliding windows to enable face detection in our implementation and refer to this as Gaussian log-likelihood with sliding windows (\textbf{GLL+SW}).}

\vspace{-0.1cm}
\subsection{Label Adaptation}
A  common evaluation approach that is especially crucial for synthetic data {and comparison of 2D and 3DA-2D landmarks} is label adaptation \cite{wood2021fake, wood20223d}. It allows us to adapt to different landmark conventions between the training dataset and an evaluation dataset. 
Given two sequences 
$(\hat{\mathbf{y}}_k, \mathbf{y}_k)_{k=1}^K$ with predicted landmarks 
$\hat{\mathbf{y}}_k \coloneqq (\hat{y}^l_k)_{l=1}^{\hat{L}} \in \mathbb R^{2 \hat L}$, 
and with ground truth landmarks 
$\mathbf{y}_k \coloneqq (y^l_k)_{l=1}^L\in \mathbb R^{2 \hat L}$ with possibly $\hat{L} \neq L$, 
we aim to train a third label adaptation model $T_\zeta: \mathbb{R}^{2 \hat{L}} \to \mathbb{R}^{2 {L}}$. 
Here, we use the loss
\begin{equation}
    \mathcal{L}(\zeta) = \frac{1}{K} \sum_{k=1}^K\| T_\zeta(\hat{\mathbf{y}}_k)  - \mathbf{y}_k\|_1.
     \label{loss_zeta}
\end{equation}
Since we deal with low-dimensional data, we use a simple multi-layer fully connected perceptron with 5 layers for this task{, see supplementary material {for details}}. 

\vspace{-0.2cm}
\section{Experiments}
\revise{We experimentally evaluate our T-FAKE dataset and the underlying model. Firstly, we assess the quality of an RGB+Thermal landmarker with integrated face detection trained on FAKE+T-FAKE in comparison to available pre-trained thermal landmarkers and state-of-the-art RGB landmarkers. 
Secondly, we quantify the perceptual T-FAKE quality based on an unsupervised metric. Thirdly, we report the temperature prediction performance of the underlying model using the mean-squared error (MSE).}
%In this section, we evaluate our generated da
%present details of our final implementation and our benchmarking pipeline. 
Note that we clamp the temperature of all thermal images between  $\SI{20}{\degreeCelsius}$ and $\SI{40}{\degreeCelsius}$ for all experiments and visualizations.

% \subsection{Thermalization Implementation}
\vspace{-0.1cm}
\subsection{Implementation Details}
\myparagraph{Thermalization.}
We train on the \mopi{RGB+Thermal SEJONG dataset recorded in lab conditions %, \textcolor{red}{the DRIVE-IN,} 
and the large-scale synthetic FAKE dataset featuring in-the-wild RGB images}. We train on a resolution of $256 \times 256$. 
{The resolution-agnostic U-net allows to} apply our two models {trained on $256\times256$ images} to the FAKE dataset at the original $512 \times 512$ resolution to generate the T-FAKE dataset. This results in the up-to-date largest thermal landmarking dataset with 200k images of {at least} 2k unique subjects and two thermal settings, see Table \ref{tab:ds}. {We refer to the supplementary for details.}

% \subsection{Landmarker Implementation}
\myparagraph{Face-Detection+Landmarker.}
We simultaneously train our landmark prediction model using a sparse 70-point {3DA-2D} and dense {478}-point {2D} landmark convention with an image size of $224 \times 224$ on our new T-FAKE thermal dataset and the original FAKE RGB dataset.  \revise{Additionally, we mix texture images 
with random landmarks 
\mopicr{as negative samples}
into the training to create meaningful uncertainty estimates $\sigma_l$.} Due to the limited, public availability of the dense {3DA-2D} ground truth in the FAKE dataset \cite{wood2021fake}, we use the {478}-point {2D} predictions of the MediaPipe model for the synthetic RGB images as our dense ground truth dense annotations.

{During inference, we use a multi-scale sliding window evaluation to generalize our model to varying image sizes without direct face tracking \revise{beyond pre-defined bounding boxes}. For our final landmark prediction, we pool all predictions and use the \revise{window with the smallest average standard deviation across all scales and all sliding windows}. \revise{To integrate face detection into the model, we use the average $\bar{\sigma}$ of the landmark uncertainties $\sigma_l$, $1\leq l \leq L$ of the final window per frame and detect faces only if $\bar{\sigma}$ is less than {a fixed threshold $t$}, i.e., GLL+SW ($\bar{\sigma} < t$). Hence, $\bar{\sigma} < \infty$ corresponds to evaluation on all frames}.  We refer to the supplementary for more details.}

 \myparagraph{Label Adaptation.}
% \subsection{Label Adaptation Implementation}
Our label adaptation network is trained solely on the two-dimensional landmark predictions and ground truth annotations of {our random} training split of a benchmark dataset and not on the images. This dataset is later excluded from the final evaluation. We do not use the predicted standard deviation at this stage. {Again, we use random rotations and shear transformations during training.}  

\myparagraph{Evaluation Metrics for Landmarking.}
We report the \textit{normalized mean error} (NME) \cite{wood2021fake, sagonas2016300, zhu2012face} for evaluation, i.e., the mean absolute error (MAE) normalized with an image-dependent constant to account for varying image sizes. Often, normalization is performed by the distance between the outermost eye landmarks  (NME IO) \cite{wood2021fake, sagonas2016300}. However, 
{the CHARLOTTE benchmark dataset} contains many {examples with side profile views where only one of the eyes is visible}. Therefore, we use the average height and width of the bounding box around the face for normalization \cite{zhu2012face} (NME W/H). 
Apart from the NME, we report the failure rates for each of the reference methods. 
The failure rate is the percentage of frames where no faces could be found. Frames without recognized faces are not included nor weighted in the NMEs. The method of rejection varies for each landmarker. We use vanilla parameters except for MediaPipe \cite{lugaresi2019mediapipe}, where we set the minimal detection confidence to $0.01$. \revise{We report T-FAKE landmarker results for different levels of face detection confidence to illustrate the trade-off between failure rate and NME.}

\begin{table*}[ht!]
\centering
%\tiny%{\small{
\resizebox{.83\textwidth}{!}{
\begin{tabular}{lllrrrrr}
Metric & Method & {Training Dataset} & {High} & {Low} & {Side} & {Front} & {Full} \\
\hline\hline
\multirow{18}{*}{NME {W/H $\downarrow$}} 
& \gray{3FabRec \cite{browatzki2020}} & \gray{300W$^\dagger$} & \gray{0.0931} & \gray{0.1084} & \gray{0.0675} & \gray{0.1344} & \gray{0.1009} \\
& \gray{3FabRec*}  & \gray{300W$^\dagger$} & \gray{0.0814} & \gray{0.1558} & \gray{0.0791} & \gray{0.1602} & \gray{0.1196} \\
& \gray{Star \cite{zhou2023star}} & \gray{300W$^\dagger$} & \gray{0.0707} & \gray{0.1021} & \gray{0.0673} & \gray{0.1064} & \gray{0.0868}\\
& \gray{Star*} & \gray{300W$^\dagger$} & {\gray{0.0639}} & \gray{0.1273} & \gray{0.0703} & \gray{0.1227} & \gray{0.0965}\\
& Mediapipe \cite{lugaresi2019mediapipe} & \textit{undisclosed$^\dagger$} & 0.1900 & 0.2745 & 0.2879 & 0.1865 & 0.2262\\
& Mediapipe* & \textit{undisclosed$^\dagger$} & 0.1400 & 0.2621 & 0.1838 & 0.2164 & 0.2023\\
& FA \cite{bulat2017far} & LS3D-W$^\dagger$ & 0.1070 & 0.2790 & 0.1222 & 0.2508 & 0.1875\\
& FA* & LS3D-W$^\dagger$ & 0.1007 & 0.2207 & 0.1075 & 0.2123 & 0.1593\\
%& \textbf{Ours} & \textbf{0.0750} & \textbf{0.1485} & \textbf{0.0403} & \textbf{0.1171} & \textbf{0.1127}\\

& DAN \cite{kopaczka2019modular} & AACHEN & 0.1677 & 0.2405 & 0.2942 & 0.1731 & 0.2054\\
& YOLO5Face \cite{kuzdeuov2022tfw} & TFW & 0.0787 & 0.1249 & 0.0701 & 0.1311 & 0.1012 \\
& U-Net+Wing \cite{kuzdeuov2022sf} &SF-TL54 & 0.2476 & 0.3134 & 0.1757 & 0.3596 & 0.2829 \\
& Dlib \cite{kuzdeuov2022sf} & SF-TL54&  0.2159 & 0.2174 & 0.1365 & 0.2741 & 0.2167 \\
%& RGB & 0.1169 & 0.3250 & 0.0667 & 0.2333 & 0.2238\\
%& Thermal & 0.0871 & 0.1487 & 0.0533 & 0.1227 & 0.1187\\
%& {Thermal Only finetune} & {T-FAKE} &  0.0832 & 0.1334 & 0.0677 & 0.1503 & 0.1090 \\
% & {Ours dense {finetune} ($\bar{\sigma} < \infty$)} & FAKE + T-FAKE & 0.0805 & 0.1285 & 0.0846 & 0.1258 & 0.1050 \\
 & GLL+SW ($\bar{\sigma} < \infty$) \cite{wood2021fake, wood20223d} & FAKE$^\dagger$ & 0.1055 & 0.2675 & 0.1241 & 0.2534 & 0.1887\\
 & GLL+SW$^*$ ($\bar{\sigma} < \infty$) & FAKE$^\dagger$ & 0.0933 & 0.2682 & 0.1312 & 0.2348 & 0.1824\\
& {GLL+SW Dense ($\bar{\sigma} < \infty$)} & FAKE$^\dagger$ + T-FAKE & 0.0854 & 0.1395 & 0.0764 & 0.1500 & 0.1132 \\

%& {Ours {finetune} ($\bar{\sigma} < \infty$)} & FAKE + T-FAKE& 0.0740 & 0.1346 & 0.0684 & 0.1420 & 0.1051 \\
%& \textcolor{red}{Ours ($\bar{\sigma} < \infty$)} & 0.0794 & 0.1312 & 0.0712 & 0.1408 & 0.1060 \\
%& \textbf{0.0732} & \textbf{0.1999} & \textbf{0.0488} & \textbf{0.1437} & \textbf{0.1383} \\
%& {Ours {finetune} ($\bar{\sigma} < 6 \times 10^{-4}$)} & FAKE + T-FAKE& 0.0729 & 0.1139 & 0.0628 & 0.1219 & 0.0923 \\
& {GLL+SW Sparse ($\bar{\sigma} < \infty$)} & FAKE$^\dagger$ + T-FAKE& 0.0683 & 0.1239 & 0.0645 & 0.1294 & 0.0969 \\
& {GLL+SW Sparse ($\bar{\sigma} < 6 \times 10^{-4}$)} & FAKE$^\dagger$ + T-FAKE& 0.0671 & 0.1118 & 0.0593 &  0.1179 &  0.0889 \\

& {GLL+SW Sparse ($\bar{\sigma} < 2.2 \times 10^{-4}$)} & FAKE$^\dagger$ + T-FAKE& \textbf{0.0661} & \textbf{0.1018} & \textbf{0.0551} &  \textbf{0.1059} & \textbf{0.0815} \\
% \arrayrulecolor{gray}\cline{2-8}

% & \textcolor{red}{Ours ($\bar{\sigma} < 1.5 \times 10^{-3}$)} & 0.0773 & 0.1129 & 0.0629 & 0.1261 & 0.0943 \\
% &\textcolor{red}{ Ours Dense ($\sigma =0$)} &  &  &  &  &  \\
% & \textcolor{red}{Ours Dense ($\sigma =X$)} &  &  &  &  &  \\
%& Ours dense & 0.0920 & 0.1840 & 0.0657 & 0.1437 & 0.1393\\
\arrayrulecolor{black}\hline
\multirow{7}{*}{\shortstack[l]{Failure\\rate (\%) {$\downarrow$}}} 
%& \textcolor{red}{3FabRec} \cite{browatzki2020} &  & & &  & \\
%& \textcolor{red}{3FabRec*}  &  &  & &  &  \\
& Mediapipe \cite{lugaresi2019mediapipe} & \textit{undisclosed$^\dagger$} & 25.05 & 46.85 & 50.09 & 22.40 & 36.25\\
& Mediapipe* & \textit{undisclosed$^\dagger$} &  12.80 & 14.07 & 24.97 & 1.92 & 13.45\\
& FA \cite{bulat2017far} & LS3D-W$^\dagger$ & 12.95 & 27.43 & 21.71 & 19.07 & 20.39\\
& FA* & LS3D-W$^\dagger$ & {0.83} & {10.45} & {4.75} & {6.80} & {5.77}\\
& DAN \cite{kopaczka2018thermal}& AACHEN& 49.32 & 48.72 & 72.85 & 25.15 & 49.01 \\
& YOLO5Face \cite{kuzdeuov2022tfw} &TFW &  13.14 & 21.55 & 19.26 & 15.66 & 17.46 \\
& U-Net+Wing / Dlib \cite{kuzdeuov2022sf} & SF-TL54 & 29.29 & 22.57 & 38.16 & 13.50 & 25.84 \\
%& {Ours {finetune} ($\bar{\sigma} < 6 \times 10^{-4}$)} & FAKE + T-FAKE& 1.63 & 16.45 & 9.15 & 9.34 & 9.25 \\
& {GLL+SW ($\bar{\sigma} < 6 \times 10^{-4}$)} & FAKE$^\dagger$ + T-FAKE& 0.98 & 10.34 &  6.94 &  4.64 & 5.79 \\
& {GLL+SW ($\bar{\sigma} < 2.2 \times 10^{-4}$)} & FAKE$^\dagger$ + T-FAKE& 6.52 & 32.76 & 23.33 & 16.68 & 20.00 \\
% & \textcolor{red}{Ours Dense ($\sigma =X$)} &  &  &  &  &  \\
\end{tabular}
}
\caption{Results on CHARLOTTE dataset {splits}. The {pre-processing stack} for RGB landmarkers \cite{flotho2021multimodal} is indicated by *. {Best NME value is in bold.} %We denote the confidence threshold with %$\bar{\sigma}$. 
\mopicr{Failure rate of GLL+SW depends on the freely chosen confidence threshold $\bar{\sigma}$.}
We include all predictions for $\bar{\sigma} < \infty${, Star} and 3FabRec resulting in {an optimal} failure rate of 0 by design. {Methods in gray perform no face detection and are thus reported with additional supervision from bounding boxes generated with the ground truth landmarks. RGB datasets are indicated by $^\dagger$.} %\anna{you do not make bold numbers that are in gray area}
% MoPi: WACV Reviewer complained about it
}
\vspace{-0.4cm}
\label{tab:results}
\end{table*}

\begin{figure}
\centering
\includegraphics[width=0.99\linewidth]{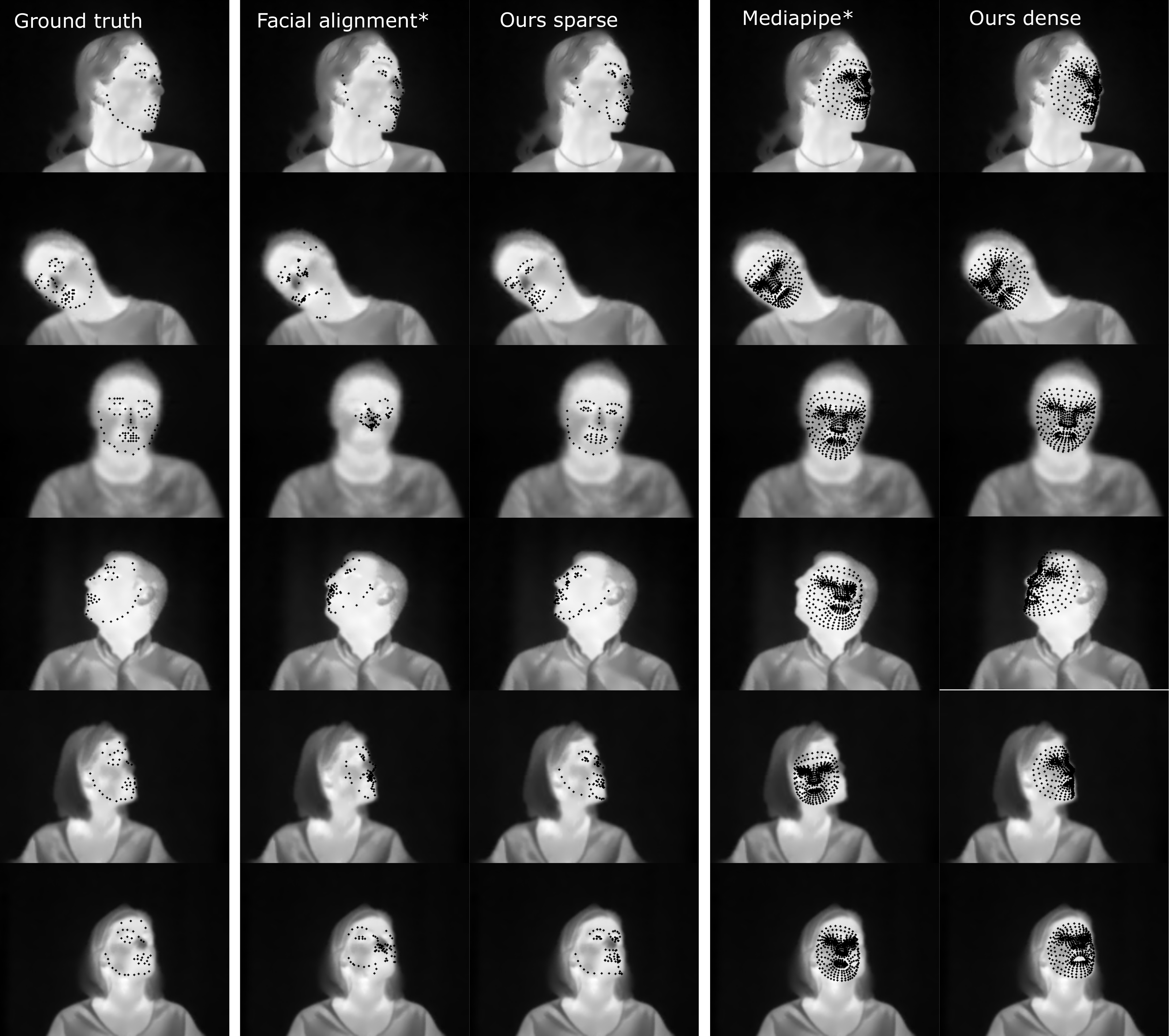}
\caption{{CHARLOTTE \cite{ashrafi2022charlotte} examples using} FA and Mediapipe with {pre-processing stack} and our models without label adaptation. While {RGB methods} can handle some challenging poses, they {may} produce broken landmarks on %certain 
{difficult} images. } 
\vspace{-0.3cm}
\label{fig:results}
\end{figure}
\vspace{-0.1cm}
\subsection{Benchmark Datasets for Landmarking}
 To compare thermal performance, we use the CHARLOTTE dataset due to its high variability in resolution, pose, and background temperature (see Figure \ref{fig:results}). The dataset is annotated with either 72 {2D} landmarks for frontal images or 43 {2D} landmarks for lateral images.  
 The 43-point landmarks represent the visible landmarks in profile pictures with our information on orientation. {We use} a total of {7684} annotated frames with different resolutions with a minimal one of around 70x60 {from the dataset}. 
For evaluation, we {propose to} split the CHARLOTTE dataset based on image size and perspective. % into high-resolution and low-resolution images. 
The {High} condition contains all images larger than the median image width of 200 pixels and {Low} all smaller images. The condition {Side} includes all images with profile annotations (43 landmarks), and {Front} contains all frontal faces (73 landmarks).
In addition, we evaluate on the established RGB landmark dataset 300W \cite{sagonas2016300} 
to verify RGB landmark prediction performance. It uses the classical 68-point convention.  {Here, we crop images to the pre-defined bounding boxes around each face for our evaluation.} {We follow the protocol from \cite{zhou2023star} for 300W evaluation.} Average image resolution is about 290 pixels.
 %It contains indoor and outdoor in-the-wild images and images with multiple faces.

%\begin{landscape}

%\end{landscape}

%\begin{table}
%\centering
%\begin{tabular}{lrr}
%  & \multicolumn{2}{c}{300W} \\
%Method & NME W/H {$\downarrow$}& NME Interocular {$\downarrow$}\\
%\hline\hline
%3FabRec \cite{browatzki2020} & 0.0376 & 0.0646 \\
%FA \cite{bulat2017far} & 0.0618 & 0.1083 \\
%{Star} \cite{zhou2023star} & & \\
%{Face Warper} \cite{liang2024generalizable} & & \\
%{Ours} & 0.0330 & 0.0577 \\%& 0.0327 & 0.0568 \\
%\multicolumn{3}{l}{Ablation Study} \\
%\hline
%{FAKE \st{RGB}} & {0.0303} & {0.0523} \\
%%RGB & 0.0344 & 0.0595
%\end{tabular}
%%\label{tab:rgb}
%\caption{Comparison of RGB performance of our method. We trained the label adaptation on \textcolor{red}{300} random samples of the combined 300W outdoor and indoor scenes %(\textcolor{red}{600\st{? or 1000?}} images) and evaluated the remaining images. \textcolor{red}{Philipp: This will completely change. We'll use the train split from STAR he%re, i.e. label adaptation on helen, lfpw, ....}}
%\vspace{-0.4cm}
%\label{tab:rgb}
%\end{table}

\begin{table}
\centering
\resizebox{\linewidth}{!}{
\begin{tabular}{lccc}
  %& \multicolumn{2}{c}{300W} \\
Model & FID {$\downarrow$} & {TUFTS MSE \cite{panetta2018comprehensive} {$\downarrow$}} & {DRIVE-IN MSE  \cite{flotho2021multimodal} {$\downarrow$}}\\
\hline
\hline 
%Model\\
%\hline
\vspace{0.05cm}
Pix2Pix \cite{isola2017image} & {.2003}  %$\pm$  .0040 
& .0421 & .1453\\
CycleGAN \cite{zhu2017unpaired} & {.2228}  %$\pm$  .0040 
& .0521 & .0730\\
CUT \cite{park2020contrastive} & {.2140}  %$\pm$  .0040 
& .0468 & .1370\\
QS-Attn \cite{hu2022qs} & {\textbf{.1173}}  %$\pm$  .0040 
& .0409 & .1303\\
\hline
$\lambda_R = 0, \lambda_W = 0$  & .5028  %$\pm$ .0039 
& {.0378} & .1117\\
$\lambda_R = 0, \lambda_W = 0.01C$ & .3146  %$\pm$  .0035 
& {.0360} & .1021\\
$\lambda_R = 1, \lambda_W = 0$ & .1706 %$\pm$ .0054 
& \underline{.0357} & \textbf{.0545}\\
$\lambda_R = 1, \lambda_W = 0.01C$ & \underline{.1598} %$\pm$ .0037 
& \textbf{.0353} & \underline{.0580}\\
\end{tabular}
}
%0.20025064800494857  - Std:  0.004019796175754065
%\label{tab:rgb}
\caption{Comparison to other models and regularizer influence on the thermalization. The parameter $\lambda_W$  corresponds to the impact of the patch-based regularizer (Eq.~\eqref{2}) and $\lambda_R$ corresponds to the segmentation-based regularizer (Eq.~\eqref{3}). We use $C=(5 \cdot 8^2)^{-1}$. The best value is in bold and the second-best is underlined.}
\vspace{-0.5cm}
\label{tab:th_ablation}
\end{table}

% \begin{table*}[ht!]
% \centering
% \begin{tabular}{llrrrrr}
% Metric & Method & {High} &  {Low} & {Side} &  {Front} & {Full} \\
% \hline
% \hline
% %\multicolumn{2}{l}{Ablation study}\\
% %\hline
% \multirow{3}{*}{NME {W/H $\downarrow$}} & RGB%+Gray 
% & 0.1055 & 0.2675 & 0.1241 & 0.2534 & 0.1887\\
% & RGB* %+Gray*
% & 0.0933 & 0.2682 & 0.1312 & 0.2348 & 0.1824\\
% %& RGB & 0.1169 & 0.3250 & 0.0667 & 0.2333 & 0.2238\\
% %& Thermal & 0.0871 & 0.1487 & 0.0533 & 0.1227 & 0.1187\\
% & Thermal Only & 0.0832 & 0.1334 & 0.0677 & 0.1503 & 0.1090 \\
% &  & - & - & - & - & - \\
% %& No label adaptation & --- & --- & --- & 0.2189 & --- \\
% \end{tabular}
% \caption{{Ablations} on CHARLOTTE {splits}. Pre-processing with the {pre-processing} stack for RGB landmarkers is indicated by *. \textcolor{black}{Here, we do not exclude low-confidence images. Hence, we report the NME on all images without failure rate.} {Best value is in bold.}}
% %Due to the \textbf{CHARLOTTE} annotation convention, only frontal images can be evaluated without label adaptation.
% \vspace{-0.2cm}
% \label{tab:results_ablation_landmark}
% \end{table*}

\vspace{-0.1cm}
\subsection{{Comparison for Landmarking}}
%State-of-the-Art Comparison for Landmarking
%To analyze how well synthetic thermal images enhance thermal landmark performance, we 
We compare the {landmarker performance with training} on our full synthetic datasets against {pre-trained} models on the CHARLOTTE dataset.  {Besides relevant} thermal {landmarkers} {we include state-of-the-art} RGB {landmarkers}. %A previous study has shown that the performance of RGB landmarkers on thermal images can be significantly boosted by applying a stack of unlearned linear filters before prediction \cite{flotho2021multimodal}. 
 {Thermal performance of RGB landmarkers can be boosted by applying a stack of unlearned linear filters before prediction \cite{flotho2021multimodal}.}
Consequently, {such pre-processing} allows for a fairer comparison between task-specific thermal landmarkers and RGB landmarkers. %Because the CHARLOTTE dataset uses its own landmark convention,  
{We use label adaptation for all methods to account for different landmark conventions for the image evaluation.} We train each label adaptation model with a 1000-image training split on CHARLOTTE similar to \cite{wood2021fake} and exclude the training split from the final evaluation. Note that this approach allows us to include our sparse model and our dense model in the evaluation. 
{To benchmark the RGB performance of the multimodal model}, we evaluate our sparse model predictions against state-of-the-art RGB sparse landmarkers on the 300W dataset. %{\st{We use label adaptation for all evaluated models with a 1000-image training split.}} %\an{it is repeated two times within one paragraph that you use 1000-image split.. }

%\subsection{Landmark Prediction Results}
%\anna{no need for a subsection + naming is not really informative - results of what? }
%\textcolor{red}{
\noindent\textbf{CHARLOTTE performance.} The NMEs and the failure rates of the different landmarkers vary drastically for the CHARLOTTE splits with generally worse performance for the Front and Low conditions. {
\revise{As for the landmarkers, we observe a clear advantage of GLL+SW ($\bar{\sigma}< \infty$) trained on FAKE+T-FAKE over GLL+SW ($\bar{\sigma}< \infty$) trained on FAKE only across sparse and dense landmarks, metrics, and splits. This confirms the suitability of our T-FAKE dataset for thermal landmarking.} The FA and Mediapipe RGB landmarkers and the SF-TL54 and Aachen thermal landmarkers show comparable performance. Using a pre-processing stack \cite{flotho2021multimodal}, leads to a drastic failure reduction for the RGB landmarkers. 
Our T-FAKE model achieves state-of-the-art performance for sparse thermal landmarking across all different conditions. The only models with comparable performance are \revise{Star} which, however, performs no face detection and is evaluated with ground truth bounding boxes, and the TFW model which, however, only predicts a set of 5 landmarks.} {While the NME of the TFW model is slightly lower than our model for $\bar{\sigma} < \infty$ for Low, Front, and Full, our model outperforms Star and TFW on almost all dataset splits in terms of NME and failure rate for $\bar{\sigma} < 6 \times 10^{-4}$.} 
%\phflo{Note that the failure rate of our method only depends on the freely chosen parameter $\bar{\sigma}$ and is zero for $\bar{\sigma}< \infty$}
These results are presented in Table \ref{tab:results} and examples are visualized in \phflo{Figure} \ref{fig:results}. \revise{The 300W-trained RGB methods Star and 3FabRec use ground truth information for tight bounding box computation. Integration of face detection into their evaluation would presumably decrease the reported NMEs.}

%\textcolor{red}{
\noindent\textbf{300W performance.} 
% To evaluate the performance of our models on RGB faces
We {evaluate the RGB performance of our multimodal model on the} 300W dataset in Table ~\ref{tab:rgb}.
% The results are presented in Table  \ref{tab:rgb}. 
% Our model, trained on both thermal and RGB faces, shows competitive performance on RGB images being surpassed only by the original model training only with RGB data. 
{Our model, trained on a combination of thermal and RGB facial images, % , demonstrates competitive performance on RGB images, being outperformed only by the original model trained exclusively on RGB data.}
{with label adaptation performs between MediaPipe and FA with label adaptation. All methods are outperformed by Star. Note that Star is the only method trained on 300W landmarks and bounding boxes. 
%Note that the bounding box has only been optimized for Star. 
Our implementation is the only one trained on general facial images and not on bounding boxes. Most importantly, training with FAKE only does not improve RGB performance.}
% \textcolor{red}{but does not reach the performance of a model trained on the original FAKE dataset.}
%While the performance of the RGB model with gray value augmentation is best, the joint model (RGB + gray + thermal) performs better than the model trained only on RGB. 
% The overall performances are consistent with the literature \cite{wood2021fake}. 

\begin{table}
\centering
\resizebox{\columnwidth}{!}{
\begin{tabular}{llrrr}
  && \multicolumn{3}{c}{300W (NME {IO $\downarrow$})} \\
Method &Training Data& Full & Comm. & Chal. \\
\hline\hline
% 3FabRec \cite{browatzki2020} &  &  & \\
% FA \cite{bulat2017far} &  &  & \\
{Star} \cite{zhou2023star} &300W& \textbf{0.0287} & \textbf{0.0252} & \textbf{0.0430}\\
%\phflo{Star} & \phflo{WFLW} & & \\
%\phflo{Star} & \phflo{COFW} & & \\ 
{\textit{Mediapipe}} \cite{lugaresi2019mediapipe} &\textit{undisclosed}& 0.0562 & 0.0469 & 0.0947 \\
{\textit{FA}} \cite{bulat2017far} &{LS3D-W}& 0.0435 & 0.0380 & 0.0661 \\
{\textit{GLL+SW ($\bar{\sigma} <\infty$)}} & FAKE & 0.0570  & 0.0494 &  0.0874 \\
{\textit{GLL+SW ($\bar{\sigma} <\infty$)}} &FAKE + T-FAKE& 0.0522  & 0.0468 &  0.0743 \\
%\multicolumn{3}{l}{\st{Ablation Study}} \\
\end{tabular}}
\caption{Comparison of RGB performance.  Label adaptation is trained on the 300W train split. Each method is evaluated on the 300W test split \cite{huang2021adnet, zhou2023star}. For Star, we use the authors' model pre-trained on 300W.  NME has been normalized with the interocular distance \cite{wood2021fake, sagonas2016300}. Methods with label adaptation are put in italic.
\phflo{Note that training with FAKE + T-FAKE does not degrade performance on 300W compared to FAKE alone.}}
\vspace{-0.3cm}
\label{tab:rgb}
\end{table}

\vspace{-0.1cm}
\subsection{Comparison for Thermalization}
\label{subsec:thermalization_comparison}
{
Our approach aims to overcome the limitations of thermal datasets to fixed camera angles, controlled lighting, and stable temperatures. However, this data limitation hinders the evaluation of `in-the-wild' images. Therefore, we report three different metrics. Firstly, we report the average \textit{Fréchet Inception Distance} (FID) \cite{heusel2017gans} between all available thermal SEJONG images and five different equally-sized thermalized FAKE subsets. The FID measures perceptual quality without ground truth image pairs. % Despite its flaws \cite{kynkaanniemi2023role, jayasumana2024rethinking}, 
Here, it compares the perceptual similarity of the thermal SEJONG dataset and thermalized FAKE datasets. Also, we report the mean-squared error (MSE) for two paired thermal datasets (with filtered background) to evaluate actual out-of-distribution temperature prediction. Here, we use TUFTS \cite{panetta2018comprehensive} and 1000 randomly selected images from DRIVE-IN \cite{flotho2021multimodal}. TUFTS features frontal lab-conditon images and DRIVE-IN contains side views from people in cars.
Table \ref{tab:th_ablation} compares our final model ($\lambda_R = 1$, $\lambda_W=0.01C$) to two supervised models trained with SEJONG only, a baseline implementation  ($\lambda_R = 0$, $\lambda_W=0$) and a Pix2Pix model \cite{isola2017image}, and three unsupervised models trained with SEJONG and FAKE, namely CycleGAN \cite{zhu2017unpaired}, CUT \cite{park2020contrastive} and QS-Attn \cite{hu2022qs}. Our model outperforms all other models for temperature prediction and displays domain adaptation for the out-of-distribution DRIVE-IN data recorded outside a lab. Moreover, the resulting T-FAKE images display better perceptual similarity than 
%images produced by 
all other methods except for QS-Attn. However, QS-Attn fails at temperature prediction, especially for the non-frontal DRIVE-IN images. 
}

\vspace{-0.1cm}
\subsection{Thermalization Ablation Study}
\label{subsec:thermalization_ablation}
{As an additional ablation of our regularizers, Table \ref{tab:th_ablation} presents FID and MSE for varied $\lambda_W$ and $\lambda_T$. The segmentation-based regularizer leads to a large boost in perceptual similarity and temperature prediction power. The patch-based regularizer leads to a smaller boost. Combining both regularizers always gives the best or second-best result confirming the effectiveness of our RGB2Thermal loss.}
\vspace{-0.2cm}
\section{Conclusion}
%{\color{red} Our results show that RGB landmark detectors can easily be refined to state-of-the-art performance on thermal images without losing accuracy on RGB benchmarks. Note that our approach can also be used for landmarking RGB images.  The insights from this work show that including imaging modalities such as thermal images in training does not limit the model's accuracy on the original domain and should motivate future synthetic datasets to include simulated thermal images or near-infrared images to enrich training.}
%\textcolor{violet}
%\an{Nope, it should be again that the thermal images are overlooked and we propose T-FAKE datasets and so on. The structure should be about the same as abstract/intro/method but with a more focus on the results or some insights}
%In this work, we focussed on facial landmarking in the thermal domain. 
Overcoming thermal data limitations, we established semi-supervised RGB2Thermal image translation. Using patch-based and segmentation-based regularization, we ensured thermalizer generalization to the synthetic FAKE dataset. This model generated the up-to-date largest thermal landmarking \revise{and segmentation} dataset, the T-FAKE dataset. Using T-FAKE, we established the up-to-date first dense thermal landmarker and a state-of-the-art RGB+Thermal sparse landmarker. We performed strict landmarker benchmarking based on established thermal and RGB methods. \revise{Moreover, the underlying RGB2Thermal model displays excellent perceptual and temperature prediction performance.} 

\clearpage
\section{Acknowledgements}
This study has partially been funded by the Federal Ministry of Education and Research (BMBF, grant numbers 13N15753 and 13N15754). M.P. acknowledges funding from the German Research Foundation (DFG) within the GRK2260 BIOQIC project 289347353. The authors acknowledge HPC resources support with hardware funded by the DFG within project 469073465. 
The authors would like to thank Daniel J. Strauss for discussions and Mayur Bhamborae and Pascal Hirsch for technical support.
{\small
\bibliographystyle{ieee_fullname}
\bibliography{egbib}
%}

\clearpage
\appendix

\section*{Supplementary Material}

{In the supplementary material, we give additional information for our method. In Section \ref{sec:supp_thermal} we provide more details on the thermalization including implementation details and an extended ablation study. In Section \ref{sec:supp_landmarking}, we add details on landmarker and label adaptation implementation. In Sections \ref{sec:supp_viz}-\ref{sec:label_adapt} we discuss limitations and provide additional result images of the datasets.}
% label_adapt
% In Sections \ref{sec:supp_viz}-\ref{sec:supp_societal} we discuss limitations, the societal impact of our work and provide additional result images of the datasets.}

\section{Thermalization}
\label{sec:supp_thermal}
\subsection{Reference Temperature Values}
As described in the main document, we train two versions of the thermalizer $T_\theta$ to {model facial temperature variations under different environmental conditions. For that purpose, we use two sets of reference temperatures, a `cold' and a `warm' condition, for the different facial regions that guide the segmentation-based regularizer. }These are presented in Table \ref{tab:temp_vals}. Thermal facial contrast is increased and overall body temperature is decreased for the `cold' condition in comparison to the `warm' condition which is in line with empirical findings \cite{ashrafi2022charlotte}.
\begin{table}[H]
\centering
\begin{tabular}{lrr}
Segmentation & `Cold' & 'Warm' \\
\hline\hline
Background, Glasses & $<$20 & $<$20\\
Skin & 33 & 35\\
Nose & 31.5 & 35\\
Eyes &34&35\\
Brows&31&34\\
Ears&32&35\\
Mouth Interior&35&35\\
Lips &32.5&35\\
Neck&34&35\\
Hair&30&30\\
Beard&31&32\\
Clothing&30&32\\
Headwear, Facewear&28&28\\
%&28&28
\end{tabular}
%\label{tab:rgb}
\caption{Reference temperatures in Celsius for our segmentation-based regularizer for the `cold' and the `warm' setup. Note that our pixel range goes from $\SI{20}{\degreeCelsius}$ to $\SI{40}{\degreeCelsius}$.}
\label{tab:temp_vals}
\end{table}

\subsection{Thermalization Implementation Details}
{For the patch-based regularizers, we use random batches with patch size of $8$. We disregard the background by excluding synthetic patches based on the ground truth background segmentation and completely black real patches. For our multi-scale approach, we sum the regularizer over $5$ scales with a downsampling factor of $0.5$. Again, facial temperatures depend on the surrounding temperature.
%, e.g., average nose temperatures of $\SI{32}{\degreeCelsius}$ at
%$\SI{20}{\degreeCelsius}$ room temperature and nose temperatures of 
%$\SI{35}{\degreeCelsius}$ at 
%$\SI{27}{\degreeCelsius}$ room temperature \cite{ashrafi2022charlotte}. 
Thus, we train a `warm' and a `cold' model with different reference temperature values for the segmentation-based regularizers.
%, see Fig. \ref{fig:gan}. %Note the inverted nose areas. 
For data augmentation, we use the same random rotations and cropping for the natural RGB and thermal images. Moreover, we apply random color changes, blurring, and shadow augmentations \cite{mazhar2021random} exclusively to the natural RGB images. Next, we apply random rotations and cropping for the synthetic RGB images. Here, we also fill holes in the original `glasses' segmentation masks showing outlines of the frame only to highlight transparent, but heat-blocking glass or plastic.
%see Fig. \ref{fig:gan} (bottom right).
Lastly, we replace the original background with a black background based on the known semantic segmentation. }
We use a U-Net $T_\theta$ with a Resnet34 encoder pre-trained on ImageNet \cite{deng2009imagenet} and train it for 10 {FAKE} epochs with SEJONG batches of size $64$ and FAKE batches of size $64$. {This corresponds to approximately 100 SEJONG epochs}. Further, we use an Adam optimizer with an initial learning rate of 0.001 which we reduce to 0.0001 after 4 epochs. Based on a random split, we use 80\% of the SEJONG data for training and all available FAKE data. Also, we use the \textit{geomloss} \cite{feydy2019interpolating} implementation for the Wasserstein patch loss with $\lambda_E = 0.1^6$. We normalize the squared error loss $\|\cdot - \cdot\|_2^2$ by dividing with the image dimension, here $256^2$. Based on the $5$ scales and patch dimension $8^2$, we set $\lambda_W = 0.01 C$ for our final model with the normalization constant $C = (5\cdot 8^2)^{-1}$. We set $\lambda_R = 1$. 
Note that the choice of $\lambda_R=1$ and $C$ are motivated to control the value range. The (normalized) MSE data fidelity term $\|\cdot - \cdot\|_2^2$ which we evaluate on normalized SEJONG images takes values in the range $[0, 1]$ for arbitrary images with pixel range $[0, 1]$ due to dimensional normalization.
To have a similar value range for the evaluated FAKE images, the patch-based regularizer $W$ takes values in $[0, 1]$ 
for $\lambda_W = C$ with arbitrary normalized images. Moreover, the segmentation-based regularizer $R$ also takes values in $[0, 1]$ for such images given arbitrary normalized reference temperature values in $[0, 1]$ for $\lambda_R=1$.

\subsection{SEJONG Dataset}
The SEJONG dataset illustrates the impact of various disguises, including glasses, wigs, and fake beards. Each subject in the dataset is presented with different disguises. As a result, it includes a lot of clothing and hairstyle variations. Most participants have a Southeast Asian or Central Asian ethnic background, but people of other ethnicities are included too. The number of participants identifying as male or female is balanced. This makes it an attractive dataset candidate for thermalization training that is supposed to generalize to a large variety of subjects. Due to reasons of data privacy, we refer to the original publication \cite{cheema2021sejong} and {have to} abstain from showing {additional} SEJONG images.
%\begin{figure}rev
%    \centering
%    \includegraphics[width=\linewidth]{Thermal/images/SejongViz.png}
%    \caption{RGB (top) and thermal (bottom) SEJONG samples.}
%    \label{fig:sejong}
%\end{figure}

%\subsection{Additional T-FAKE Samples}
%Figure \ref{fig:sample_tfake} shows more uncurated T-FAKE samples with the `cold' and `warm' variants.

{
\subsection{Generalization to Out-of-Lab Conditions}
The success of neural networks in the last two decades has been tremendous, especially in the imaging domain. Nevertheless, a common empirical finding is the limited capability of neural networks to generalize to new settings \cite{novak2018sensitivity}. Often, this limitation is caused by biased datasets. Particularly in the biomedical domain, data acquisition is a valuable task. However, due to real-world restrictions, data is often acquired in laboratory conditions. Acquiring paired RGB and thermal facial images requires a calibrated multimodal camera setup. Most multimodal facial datasets are restricted to frontal views with relatively neural expressions and frontal lighting, e.g. \cite{panetta2018comprehensive, cheema2021sejong, flotho2021multimodal}. As a direct consequence, most Thermal2RGB research has focused on learning the transformation purely for \textit{frontal images with frontal lighting at room temperature} \cite{riggan2018thermal, zhang2018tv, nair2023t2v}. To our knowledge, our regularized model is the first facial model to promote the explicit generalization of a learned thermal transformation to new poses, facial expressions, and lighting conditions and to simulate distinct temperature conditions. However, due to a lack of paired multimodal `in-the-wild' facial datasets, we have to partially resort to metrics for unsupervised image translation, i.e., the FID \cite{heusel2017gans}. To additionally visualize this result, we show results for the TUFTS \cite{panetta2018comprehensive} dataset containing RGB and thermal images. Again, we find that paired RGB and images are only available for frontal views. We present the results of applying our baseline model without regularization ($\lambda_W = 0$, $\lambda_R=0$), a Pix2Pix model, and our final `cold' model to the dataset subset with paired RGB images in Figure \ref{fig:tufts}. We chose the `cold' model because the images were recorded at room temperature. We see only a marginal impact of our regularization on the prediction. However, the TUFTS dataset additionally contains RGB images without paired thermal images recorded from different angles. Therefore, we also apply both models to random images taken from a fixed side angle. The result is displayed in Figure \ref{fig:tufts_side}.
We see that the unregularized model generates various large facial artifacts whereas our final model contains almost no facial artifacts. This shows that the main limitation of training RGB2Thermal and Thermal2RGB models is the lack of paired `in-the-wild' multimodal images. Our regularization allows us to overcome this limitation.
}
\begin{figure}%[H]
    \centering
    \includegraphics[width=\linewidth]{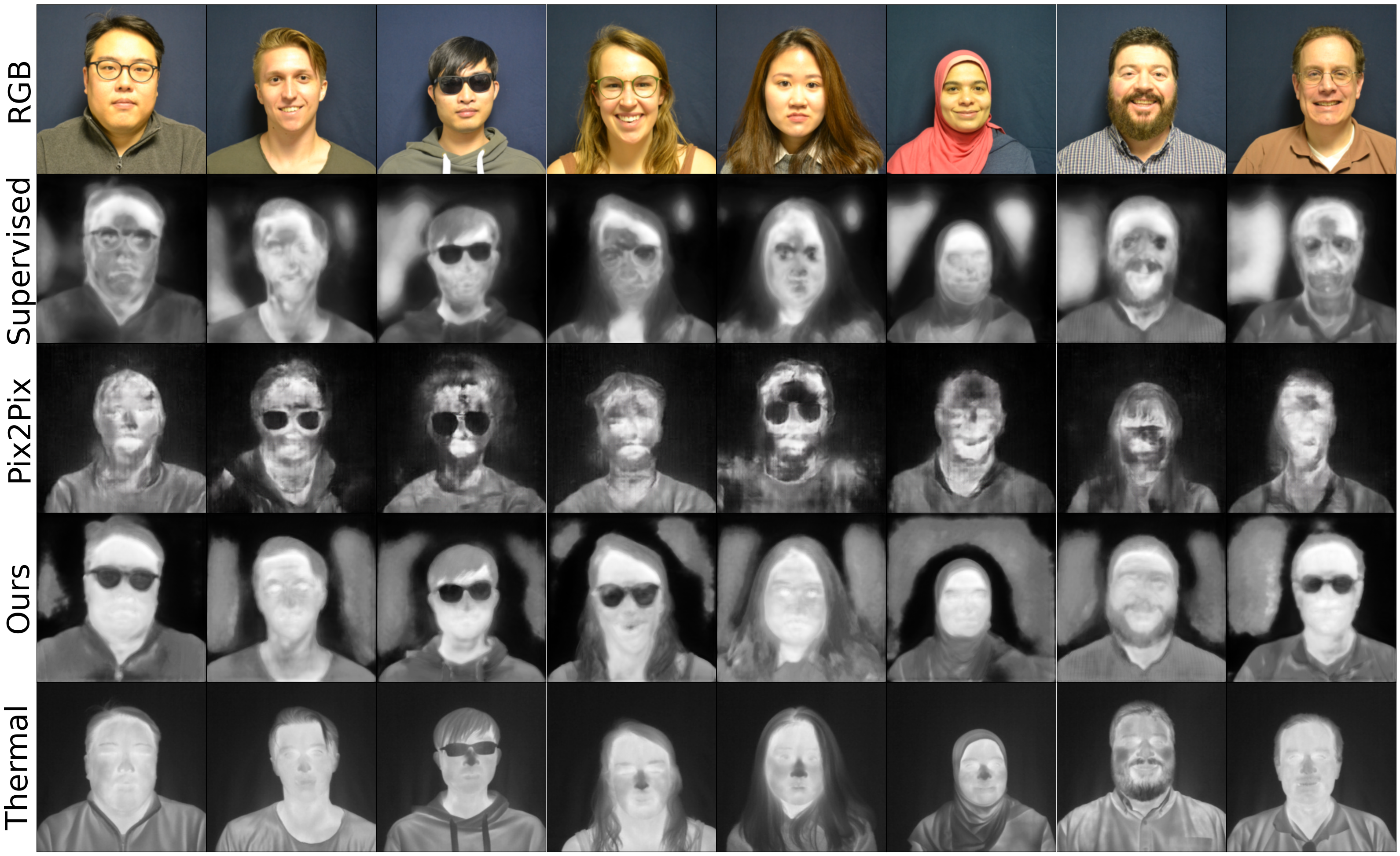}
    \caption{Thermalization results for \textit{frontal} RGB images \textit{with} paired thermal images from the TUFTS \cite{panetta2018comprehensive} database without regularization ($\lambda_W = 0$, $\lambda_R=0$), Pix2Pix and our final model ($\lambda_W = 0.001C$, $\lambda_R=1$) (top to bottom).}
    \label{fig:tufts}
\end{figure}

\begin{figure}%[H]
    \centering
    \includegraphics[width=\linewidth]{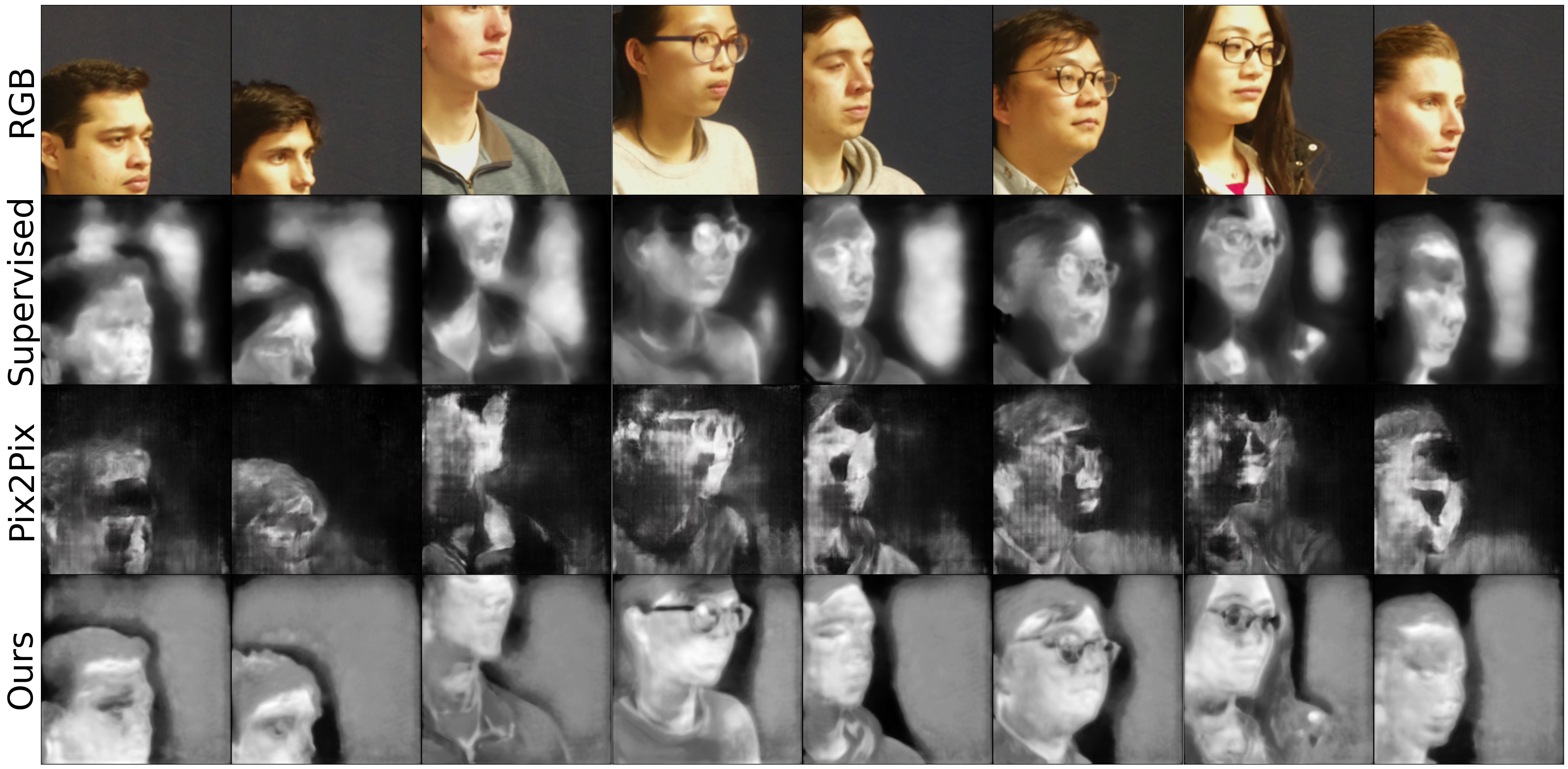}
    \caption{Thermalization results for \textit{side} RGB images \textit{without} paired thermal images from the TUFTS \cite{panetta2018comprehensive} database without regularization ($\lambda_W = 0$, $\lambda_R=0$), Pix2Pix and our final model ($\lambda_W = 0.001C$, $\lambda_R=1$) (top to bottom).}
    \label{fig:tufts_side}
\end{figure}

\subsection{Thermalization Comparison Details}
{We used the official PyTorch implementation for all compared models. We trained Pix2Pix and all SEJONG images and all unsupervised images using all $\sim$10k SEJONG images and 10k FAKE (RGB) images. We used default hyperparameters and trained all models for 10 epochs with a constant learning rate and an additional 10 epochs with a linearly decreasing learning rate. All models were trained and evaluated with a resolution of $256$ for the FID and the MSE. However, our models were evaluated on a resolution of 512 and the output was downscaled to 256 to be in line with the final T-FAKE dataset. We display examples in Figure \ref{fig:sample_fake_comp}. For the MSE comparison, we removed the background because we use random background augmentations for our final landmarker training. Here, we removed the background by masking all predictions based on a $\SI{21}{\degreeCelsius}$ threshold based on the ground truth. For the FID evaluation, we randomly choose the `warm' or the `cold' variant for each image for our model. For the MSE, we average the results for `warm' and `cold' variants. Due to data privacy reasons, we are only able to present ground truth thermal images for two persons from the DRIVE-IN dataset \cite{flotho2021multimodal}, see Figure \ref{fig:drive}.}

\begin{figure}%[H]
    \centering
    \includegraphics[width=\linewidth]{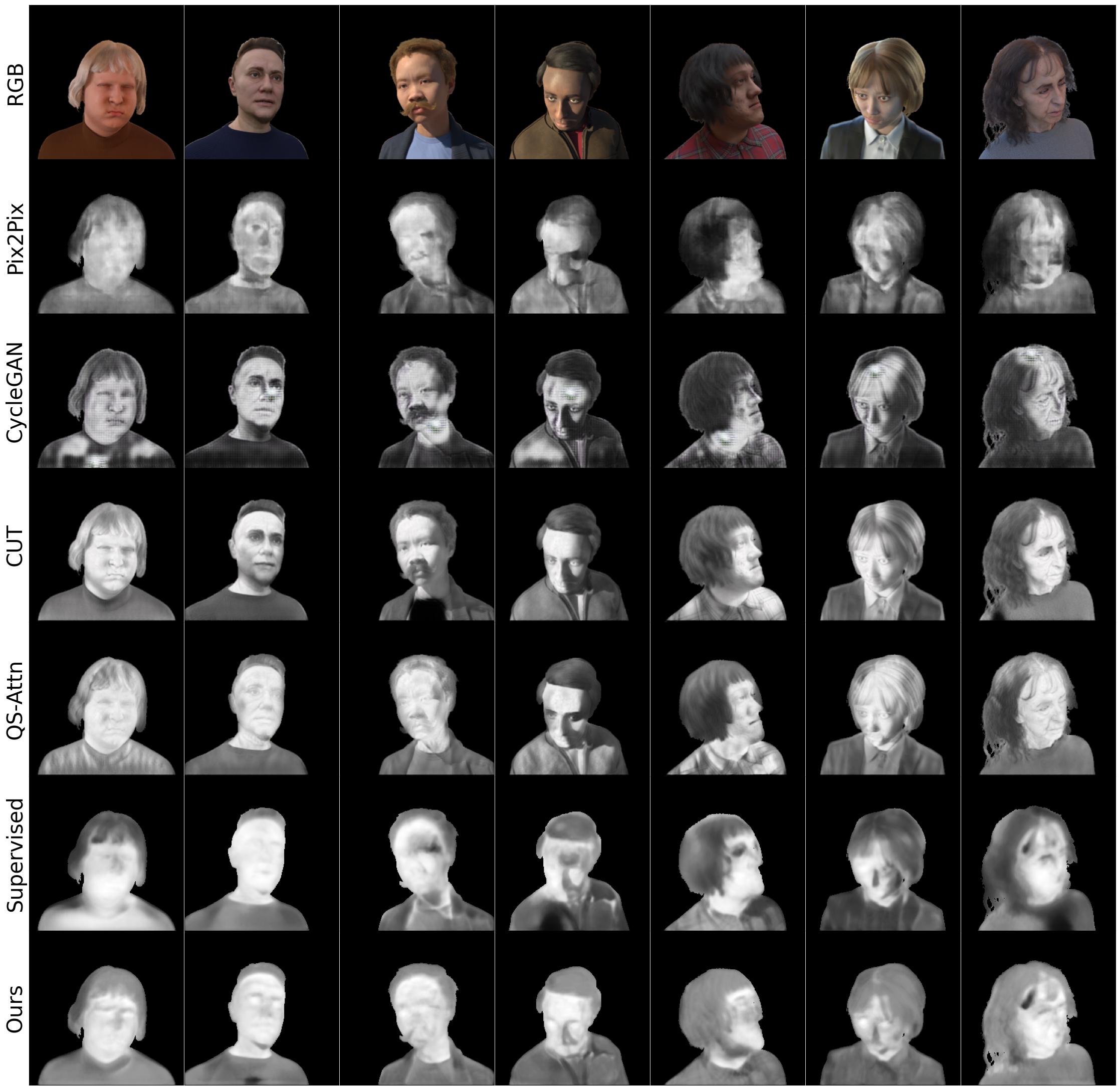}
    \caption{Comparison of FAKE images thermalized with Pix2Pix \cite{isola2017image}, CycleGAN \cite{zhu2017unpaired}, CUT \cite{park2020contrastive}, QS-Attn \cite{hu2022qs}, our supervised baseline and our final model (top to bottom, background removed).}
    \label{fig:sample_fake_comp}
\end{figure}

\begin{figure}%[H]
    \centering
    \includegraphics[width=\linewidth]{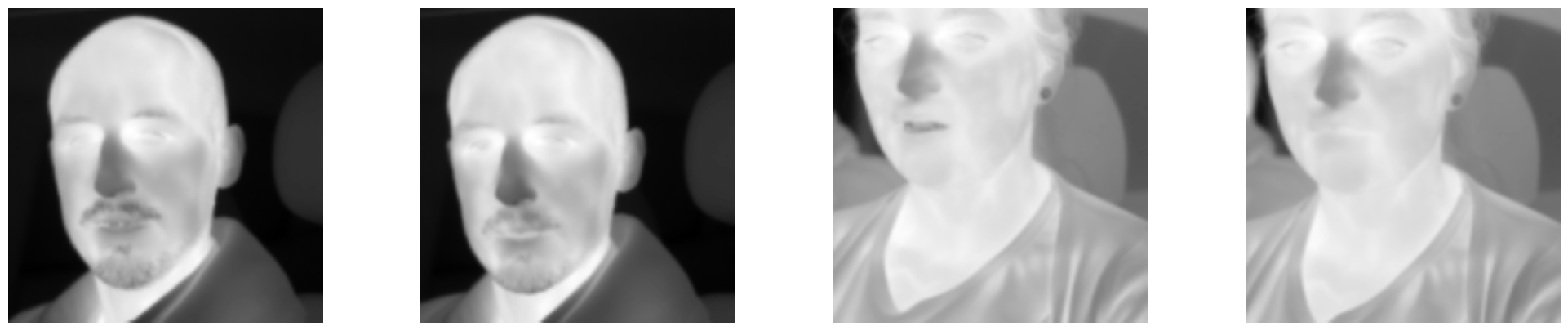}
    \caption{Thermal ground truth samples from the DRIVE-IN \cite{flotho2021multimodal} dataset with side profile.}
    \label{fig:drive}
\end{figure}

\subsection{Extended Thermalization Ablation Study}
\begin{figure}%[H]
    \centering
    \includegraphics[width=\linewidth]{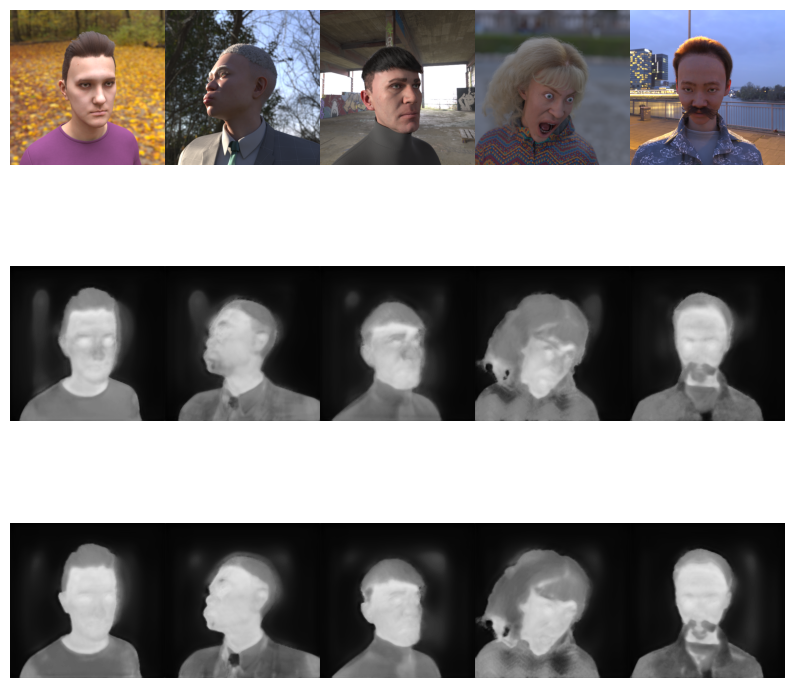}
    \caption{T-Fake samples with original images (first row), `cold' images' (second row), and `warm' images' (third row). Note the diminished contrasts of the noses and the checks.}
    \label{fig:sample_tfake}
\end{figure}
Given that the regularizers are solely defined for the synthetic images, we fix $\lambda_T = 1$ to ensure that the regularizer is on the same scale as the MSE of the real images and vary $\lambda_W$ for our ablation. We train all models with the same setup and random seeds. The FID implementation is the default \textit{PyTorch-Ignite} \cite{ignite} implementation. Here, we extend the ablation study table in the main document which only displays the best result for $\lambda_W = 0.01C$. In particular, we display the results of a grid search for $\lambda_W$ and $\lambda_R$ in Table \ref{tab:th_ablation_full}. As described in the main document, we calculate the mean FID and its standard deviation for {five} different subsets of the T-FAKE dataset. Moreover, we use the same setup to compare the perceptual quality of the `cold' and the `warm' setup of our final T-FAKE dataset, see Table \ref{tab:ablation_warm_cold}. Figure \ref{fig:sample_tfake} shows T-FAKE samples with the `cold' and `warm' variants. In addition, we present some samples generated with different regularization configurations in Figure \ref{fig:cold_ablation} for the `cold' setup and in Figure \ref{fig:warm_ablation} for the `warm' setup. Here, we can visually see the impact of the different regularizers. 
\begin{comment}
    {In addition, we display the MSE after each FAKE epoch on a 20\% test split of the SEJONG dataset for final models and the fully supervised baseline model ($\lambda_W = 0$, $\lambda_R=0$). We can see that the fully supervised model largely converges and that it consistently outperforms our final models on SEJONG test cases. Nevertheless, Figure \ref{fig:cold_ablation}, Figure \ref{fig:warm_ablation}, and Table \ref{tab:ablation_warm_cold} show that despite its good SEJONG performance the supervised model fails to generalize to the FAKE dataset and that our regularization enables generalization.}
\end{comment}

\begin{table}%[H]
\centering
\begin{tabular}{lrr}
  %& \multicolumn{2}{c}{300W} \\
Regularization & $\lambda_R =1$ & $\lambda_R =0$ \\
\hline\hline
$\lambda_W= 1C$  & .1665  $\pm$ .0030&  .3375  $\pm$ .0056\\
$\lambda_W= 0.1C$& .1753 $\pm$  .0018& .3654  $\pm$  .0032\\
$\lambda_W= 0.01C$ &  .1598 $\pm$  .0041& .3146 $\pm$ .0092\\
$\lambda_W= 0$ & .1706 $\pm$ .0029& .5028 $\pm$ .0054\\
%\textbf{Both} & 0.2719227260493127  - Std:  0.005430191098617782&
\end{tabular}
%\label{tab:rgb}
\caption{Impact of the regularization parameters on the perceptual quality measured with the FID {($\downarrow$)} with $C = (5\cdot 8^2)^{-1}$.}
\label{tab:th_ablation_full}
\end{table}

\begin{comment}
begin{figure}%[H]
    \centering
    \includegraphics[width=\linewidth]{Thermal/images/Test_Loss.png}
    \caption{MSE {($\downarrow$)} on SEJONG test split after different numbers of FAKE epochs for our final `cold' and `warm' model ($\lambda_W= 0.01C, \lambda_R = 1$) and our fully supervised baseline model ($\lambda_W = 0$, $\lambda_R=0$)}
    \label{fig:test_loss}
\end{figure}
\end{comment}
According to the FID, both regularizers have a positive impact. The segmentation-based regularizer greatly boosts the perceptual quality, while the effect of the patch-based regularizer is smaller. The optimal FID value is obtained for $\lambda_W=0.01C$ and $\lambda_T=1$, the parameters of our final model. The perceptual quality of the figures seems in line with the FID.  A closer look at the last two rows in both figures shows that the segmentation-based regularizer alone can lead to smoothed facial areas and overly exaggerated differences on the edges of the semantic segmentation. This becomes more apparent for the `cold' setup as it leads to more thermal contrast within the face, see Figure \ref{fig:cold_ablation}. The FID shows only a small difference between the `cold' and `warm' variants. The `warm' variant displays slightly lower FID values. For a visual comparison of the T-FAKE images with real thermal images, we refer to Fig. \ref{fig:charlotte_samples}. 

\begin{comment}
    {Moreover, we display visual thermalization results of our baseline model without regularization, and our final models on random lab-condition RGB images of the TUFTS \cite{panetta2018comprehensive} dataset. Here, we see that both models largely achieve thermalization of TUFTS images without artifacts. This observation stands in contrast to the artifacts produced by the fully supervised baseline for FAKE images. Again, this highlights the inability of a fully unsupervised model to generalize to the FAKE dataset.}
\end{comment}

\begin{table}%[H]
\centering
\begin{tabular}{lrr}
  %& \multicolumn{2}{c}{300W} \\
Setup & `Cold' & `Warm'\\
\hline\hline
FID  {$\downarrow$} & .1577 $\pm$ .0024&  .1685  $\pm$ .0111\\
\end{tabular}
%\label{tab:rgb}
\caption{Perceptual comparison of thermal setups `cold' and `warm' using the FID.}
\label{tab:ablation_warm_cold}
\end{table}

%\subsection{Comparison of `Cold' and `Warm' Setup}

\begin{figure}%[H]
    \centering
    \includegraphics[width=\linewidth]{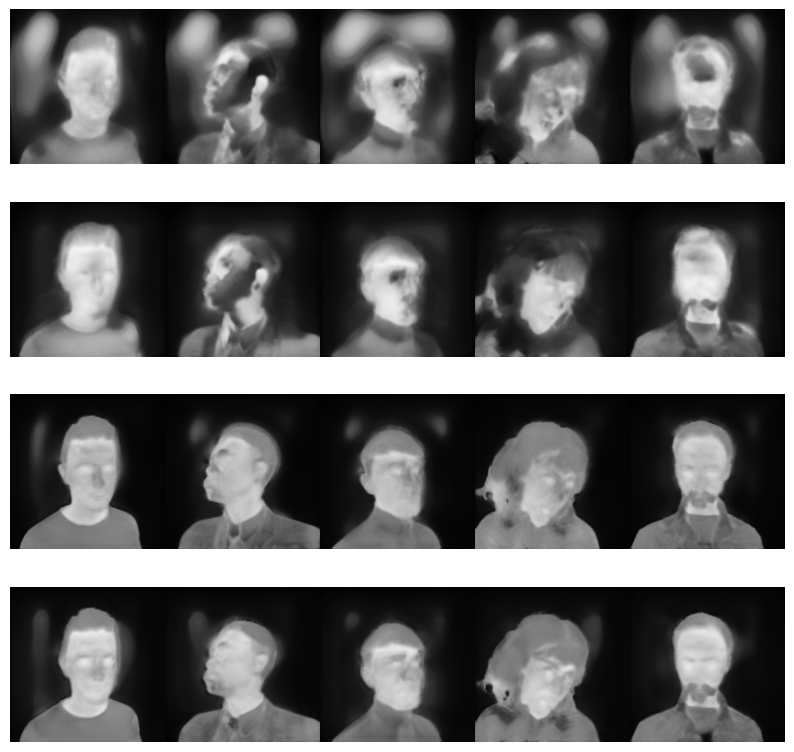}
    \caption{Regularization impact on images for `cold' setup: No regularization ($\lambda_W = 0$, $\lambda_R=0$), only patch-based ($\lambda_W = 0.01C$, $\lambda_R=0$), only segmentation-based ($\lambda_W = 0$, $\lambda_R=1$), and final model ($\lambda_W = 0.01C$, $\lambda_R=1$) (top to bottom).}
    \label{fig:cold_ablation}
\end{figure}

\begin{figure}%[H]
    \centering
    \includegraphics[width=\linewidth]{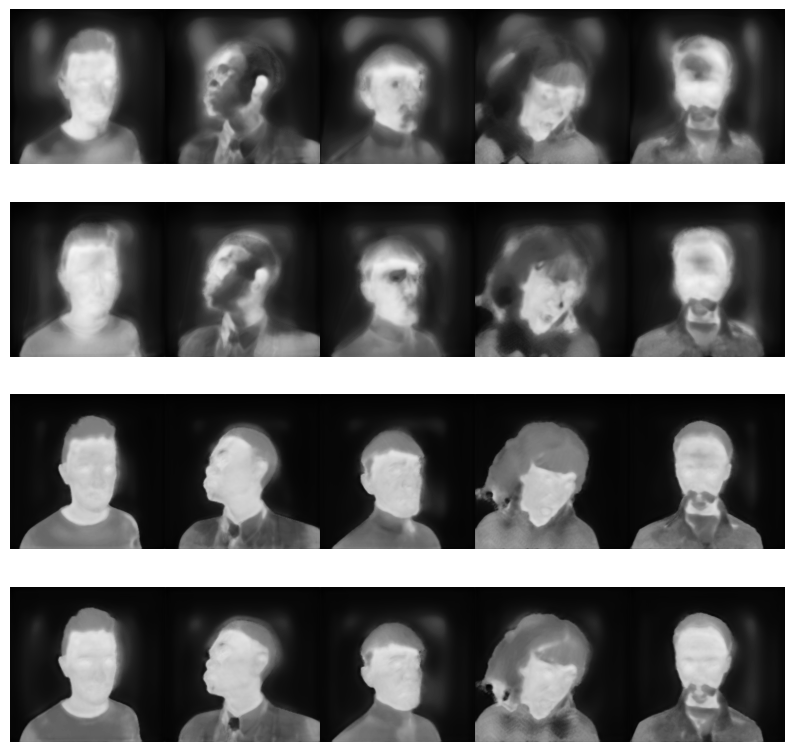}
    \caption{Regularization impact on images for `warm' setup: No regularization ($\lambda_W = 0$, $\lambda_R=0$), only patch-based ($\lambda_W = 0.01C$, $\lambda_R=0$), only segmentation-based ($\lambda_W = 0$, $\lambda_R=1$), and final model ($\lambda_W = 0.01C$, $\lambda_R=1$) (top to bottom).}
    \label{fig:warm_ablation}
\end{figure}

\begin{table*}[ht!]
\centering
%\tiny%{\small{
\resizebox{\textwidth}{!}{
\begin{tabular}{llrrrrrr}
Metric & Method & Training Dataset & High & Low & Side & Front & Full \\
\hline\hline
\multirow{4}{*}{NME W/H $\downarrow$} & GLL + RW ($\bar{\sigma} < \infty$) \cite{wood2021fake, wood20223d} & FAKE & 0.1055 & 0.2675 & 0.1241 & 0.2534 & 0.1887\\
& GLL + RW* ($\bar{\sigma} < \infty$) & FAKE & 0.0933 & 0.2682 & 0.1312 & 0.2348 & 0.1824\\
& GLL + RW ($\bar{\sigma} < \infty$) & T-FAKE &  0.0832 & 0.1334 & 0.0677 & 0.1503 & 0.1090 \\
& GLL + RW ($\bar{\sigma} < \infty$) & FAKE + T-FAKE& 0.0740 & 0.1346 & 0.0684 & 0.1420 & 0.1051 \\
\end{tabular}
}
\caption{Ablation results on CHARLOTTE dataset splits. Pre-processing with the pre-processing stack for RGB landmarkers is indicated by *. The confidence threshold has been set to infinity. RGB + Thermal (FAKE + T-FAKE) and Thermal Only (T-FAKE) models have been finetuned from the FAKE models. }
\vspace{-0.4cm}
\label{tab:results_ablation_landmark}
\end{table*}

\section{Landmarking}
\label{sec:supp_landmarking}
\subsection{CHARLOTTE Dataset}
 The CHARLOTTE dataset contains thermal images with varying thermal conditions,  various head positions, and multiple camera distances. Moreover, it contains information about the thermal sensation of the subjects. We refer to Fig. \ref{fig:charlotte_samples} for a visualization of some thermal CHARLOTTE images without landmarks.

\begin{figure}%[H]
\centering
\includegraphics[width=\linewidth]{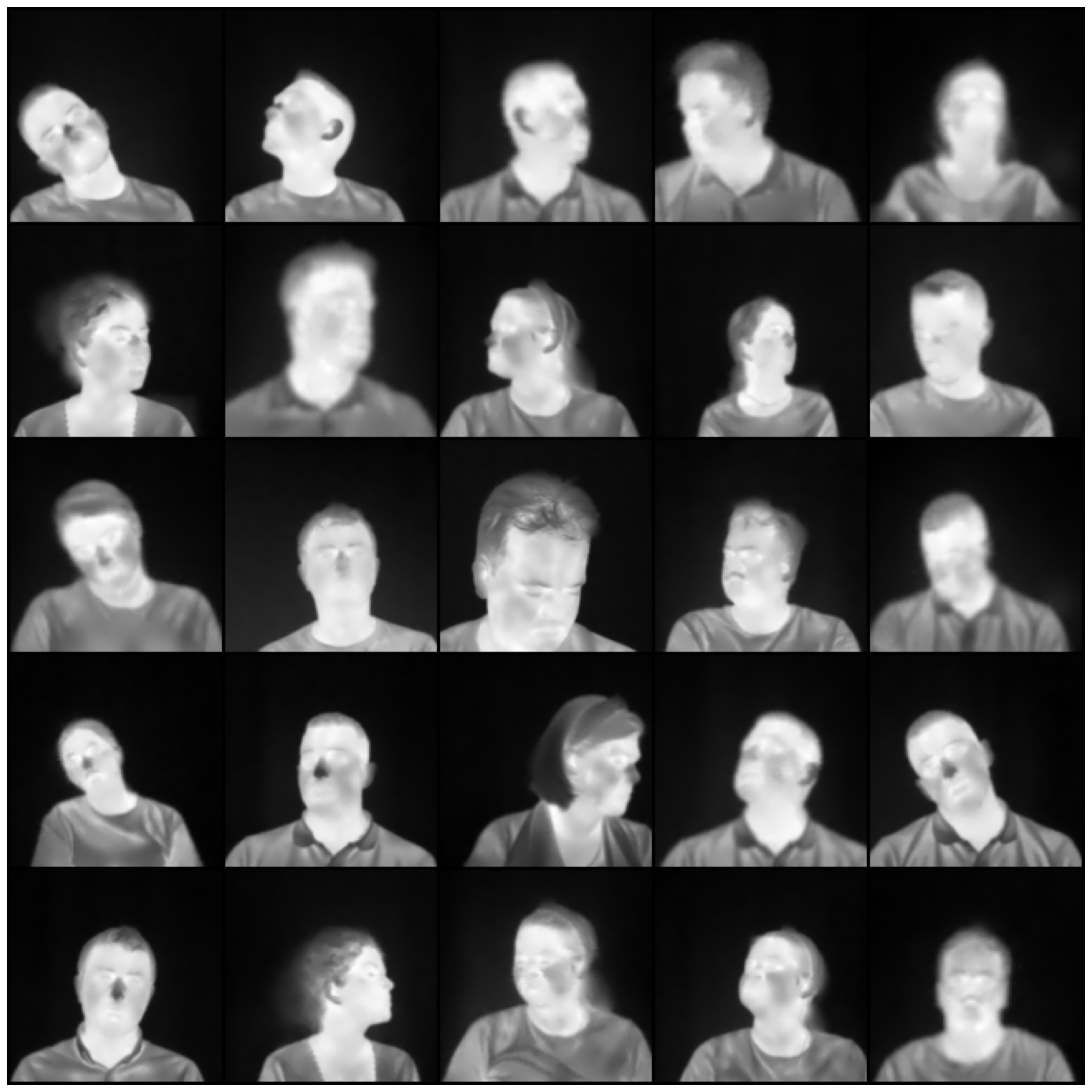}
\caption{CHARLOTTE image samples with different resolutions, environmental conditions, and subjects. }
\label{fig:charlotte_samples}
\end{figure}

\subsection{Landmarking Implementation Details}
{For training, we include random landmark positions on a texture dataset \cite{godi2019texel} as negative examples into the dataset to increase the learned uncertainty $\sigma^2$ on images without faces. %\textcolor{red}
 %\textcolor{red}
%Furthermore, we employ Gaussian blur, color jitter, and global pixel offset for thermal images during training. 
On thermal images, we fill in the background with random textures from the texture dataset \cite{godi2019texel} with a probability of 0.25. %Note that our landmark prediction model is trained with an image size of $224 \times 224$. 
During inference, we use a multi-scale sliding window evaluation to generalize our model to varying image sizes and face scales. 
%This means that we resize all images to size $465 \times 350$ and then 
We downsample iteratively with a factor of $0.75$ until the height or the width reaches $224$. For each image scale, we run our model on sliding windows of size $224 \times 224$ with a stride of $20$. For our final landmark prediction, we pool all predictions and use the landmark with the smallest predicted standard deviation across all scales and all sliding windows.}
For training the landmarker $T_\psi$, we finetune a model that has been pre-trained on the original FAKE dataset for the sparse landmarker {for the results in Table \ref{tab:results_ablation_landmark}}. During refinement, thermal images are used with a probability of 0.4 split with equal probability for `cold' and `warm' conditions. We finetune for 100 epochs with a learning rate of 0.0004, Adam optimizer with weight decay, batchsize of 512 and OneCycleLR scheduler. {Our final model is trained for 4000 epochs, a learning rate of 0.001 on FAKE and T-FAKE (p=0.4)}. 

{\noindent{\textbf{Augmentations details.}} We use of spatial augmentations, including random shear, rotations, resizing, and cropping to allow the landmarker to learn a large variation of face orientations without the need for a dedicated face detector. Specifically, we use geometric augmentations which apply random rotations (up to ±45°) and random shear (±7°) to both the image and landmarks, preserving geometric consistency. In addition, a random resized cropping operation is performed, with the cropped region size scaling between $40\%$ and $200\%$ of the original image and an aspect ratio ranging between $\frac{3}{4}$ and $\frac{4}{3}$. Furthermore, we use photometric transformations and random Gaussian smoothing. The thermal images are randomly jittered to simulate temperature variations and thermal images are randomly inverted with a probability of 0.1. Additionally, random noise is applied with a probability of 0.2 to simulate sensor noise. See Figure \ref{fig:augmentations} for image examples with augmentations applied.}

%\noindent{\textbf{Augmentations.}
%}
%\textcolor{red}{Move here by Moritz.}
\subsection{Label Adaptation Implementation Details}
Label Adaptation was trained for each method on the predictions on all detected faces on a random 1000 image CHARLOTTE split. We train a model $T_\zeta$ for 2000 epochs with a learning rate 0.002, OneCycleLR and Adam optimizer. As landmark augmentations, we apply random rotation up to 45° as well as random shearing during training. The label adaptation network is a  five-layer perceptron with fully connected layers that takes the predicted landmarks together with the resize factor as input. The latter accounts for varying degrees of quantization at different image sizes in the CHARLOTTE ground truth. 
The label adaptation generally handles even outlier predictions but can also contain fail cases, (see Figure \ref{fig:sup_adapt}).

\noindent{\textbf{RGB Model Inference.}} We use two different pre-processing approaches to include landmarkers solely developed for RGB images into the evaluation. Firstly, gray-value images, where the temperature between 20° and 45° is normalized and, secondly the pre-processing stack proposed in \cite{flotho2021multimodal}. The pre-processing stack consists of temperature clamping between 20°C and 45°C, unsharp masking with two sets of parameters with and without temperature inversion. The reported landmarks are the averages over all detected faces. This simple pre-processing stack is a simple method for boosting RGB landmarker performance for thermal images \cite{flotho2021multimodal}. As a result, we can include RGB landmarkers as a baseline for thermal landmarking models.

%\begin{comment}
\subsection{{Landmarking Ablation Study}}
{To study the impact of our thermal data, we report the CHARLOTTE results of our landmarker trained with i) RGB images only, ii) finetuned with thermal images and iii) finetuned with both FAKE and T-FAKE. Again, we use label adaptation for all variations. Table \ref{tab:results_ablation_landmark} shows the results. Training with the T-FAKE dataset significantly improves the accuracy of thermal landmarking across all conditions. In addition, multimodal training with the FAKE and T-FAKE datasets leads to better thermal landmarking performance than finetuning only with the T-FAKE dataset.}
% Moreover, we ablate our RGB performance on 300W by testing the model trained on RGB images only, see Table \ref{tab:rgb}. Here, we see a slight drop in performance caused by multimodal training. Nevertheless, our multimodal model still outperforms FA \cite{bulat2017far} and 3FabRec \cite{browatzki2020}.
%\end{comment}

\subsection{Inference Ablation}
 We analyze the impact of our inference strategy, see Table \ref{tab:window_results}. We compare inference on i.) the complete image scaled to $224 \times 224$ (whole image), ii.) followed by refinement on a bounding box computed from the predictions obtained with i.) and finally iii.) with the sliding window approach described in the main document. Method iii.) produces the best results over all images except for Charlotte low, where ii.) performs slightly better while at the same time being also suitable for real time estimation. 
 
\begin{figure*}
\includegraphics[width=\linewidth]{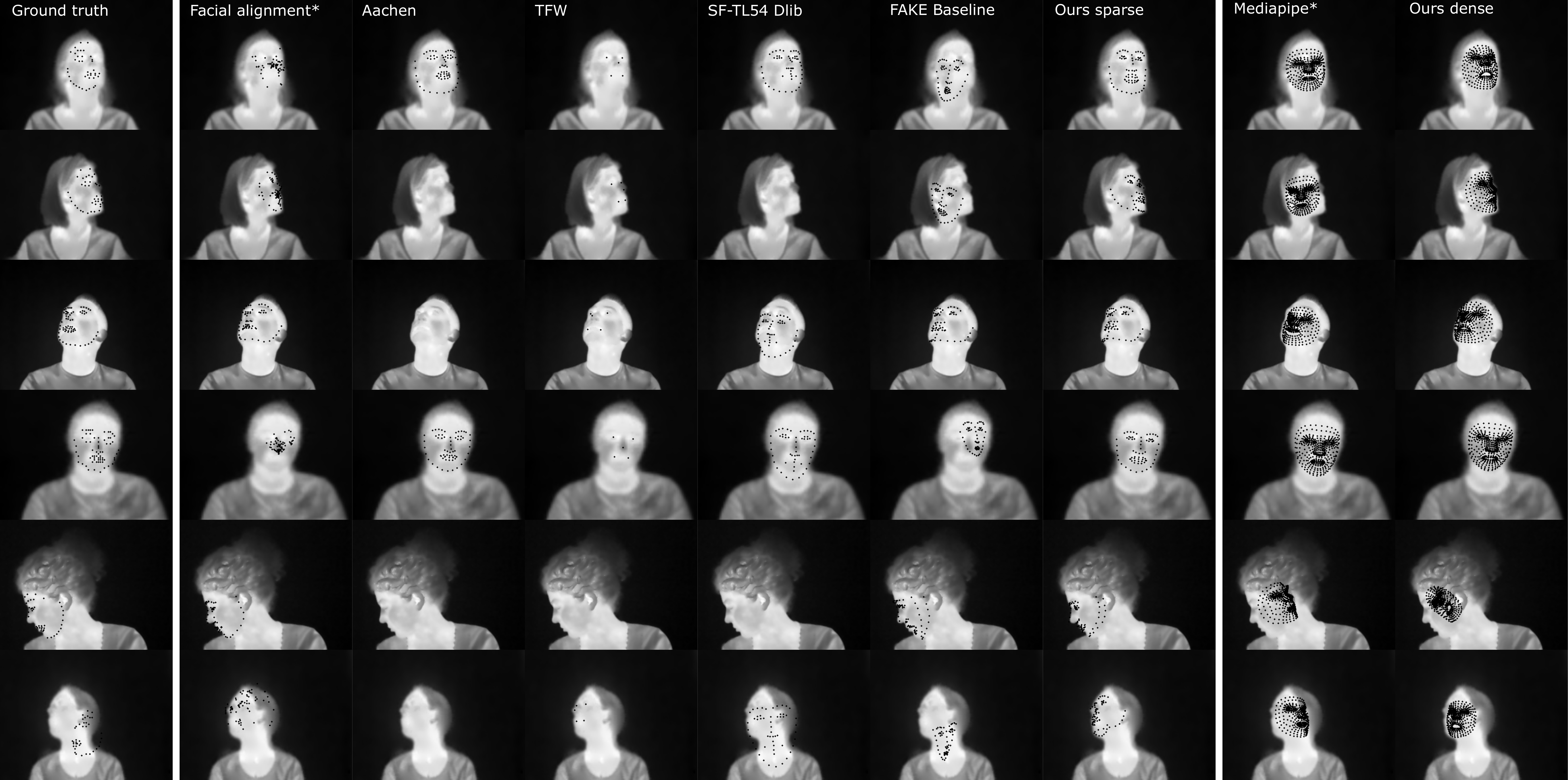}
\caption{Results on examples from the \textbf{CHARLOTTE} dataset \cite{ashrafi2022charlotte} with different RGB and thermal predictors and our models. For images without landmarks, no faces were detected. The performance of RGB methods can be greatly improved when the images are inverted or sharpened indicated by *. The first column shows the limitations in the CHARLOTTE ground truth: profile annotation convention for frontal views (1st row), quantization artefacts for low resolution images (4th row), translated annotations (last row). }
\label{fig:sup_landmarks}
\end{figure*}

\begin{table*}[ht!]
\centering
{\small{
\begin{tabular}{l|r|rrrrrr}
Method & {Inference Time (ms)} &{High} & {Low} & {Side} & {Front} & {Full} \\
\hline\hline
{Sliding window ($\bar{\sigma} < \infty$)} & {88.73} & 0.0740 & 0.1346 & 0.0684 & 0.1420 & 0.1051 \\
%{Ours dense ($\bar{\sigma} < \infty$)} & 0.0805 & 0.1285 & 0.0846 & 0.1258 & 0.1050 \\
{Refined bounding box ($\bar{\sigma} < \infty$)} & {10.41} & 0.0847 & 0.1329 & 0.0696 & 0.1494 & 0.1095 \\
%{Ours dense ($\bar{\sigma} < \infty$)} & 0.0805 & 0.1285 & 0.0846 & 0.1258 & 0.1050 \\
{Whole image ($\bar{\sigma} < \infty$)} & {5.92} & 0.1091 & 0.1868 & 0.0842 & 0.2139 & 0.1490 \\
%{Ours dense ($\bar{\sigma} < \infty$)} & 0.0805 & 0.1285 & 0.0846 & 0.1258 & 0.1050 \\
\end{tabular}
}}
\caption{{NME (W/H) ($\downarrow$) for CHARLOTTE splits with} different strategies for landmark computation. The final results are estimated with sliding windows similar to \cite{wood20223d}, however, we achieved comparable results when computing the landmarks on input images rescaled to $224\times224$. Here, we do not exclude high-uncertainty images and evaluate all images without failure rate, i.e., $\bar{\sigma} < \infty$. {Average inference time per frame on the Full split has been benchmarked on a single NVIDIA H100 80GB GPU with a batch size of 1.}}
\vspace{-0.4cm}
\label{tab:window_results}
\end{table*}

%\begin{figure}%[H]
%\includegraphics[width=\linewidth]{images/adapt/labels}
%\caption{  }
%\label{fig:labels}
%\end{figure}
%\subsection{Additional Landmarking Samples}
%\section{}

\section{Large-Scale Visualization}
\label{sec:supp_viz}
For a large number of T-FAKE samples, we refer to Figures \ref{fig:large} and \ref{fig:large_sparselandmarks}. Here, we simply use the first 128 images based on the numerical naming convention of the original FAKE dataset.
\section{Thermal Semantic Segmentation Dataset}
\label{sec:supp_segmentation}
%\phflorev{This is a very short section, should we maybe put it under limitations, as it is kind of related to future work?}
Note that by design detailed segmentation masks are available for all T-FAKE images. While the training of a semantic segmentation model was out-of-scope for our work, we want to highlight the fact that our dataset can also be used for such training. The possibility of such segmentation training with synthetic data has already been demonstrated by Wood et al. \cite{wood2021fake}.
\section{Limitations}
\label{sec:supp_limitations}
%\anna{if space is needed, this paragraph can easily go to supmat}
%\textcolor{violet}
\noindent{\textbf{Thermalization.}{
This work depends on the thermalization of the final renders in the FAKE dataset. The dataset contains very difficult lighting conditions and scene compositions that make it powerful to train landmarkers but also made the thermalization particularly challenging and could only be solved with advanced {domain-adaptive semi-supervised} regularization approaches. Despite a good perceptual result {of the faces}, some T-FAKE images {can} contain {minor} artifacts {on clothing and on the background (e.g. see Figure \mbox{\ref{fig:warm_ablation}}, bottom second from right)}. Nevertheless, these artifacts remain limited. {Moreover, background artifacts can easily be removed by choosing a suitable background based on the ground truth segmentation as implemented during our landmarker training, see Fig. \ref{fig:augmentations}. 
%In addition, our resulting images suffer from a blur error commonly associated with the MSE loss \cite{zhao2016loss, isola2017image}
%due to the simple MSE loss choice for our supervised image translation training. In our RGB2Thermal translation contribution, we focussed on the (demonstrated) improvement of out-of-distribution generalization beyond the available lab-recorded datasets. 
We merely used an MSE loss for our supervised training. Including an adversarial \cite{isola2017image, zhang2018tv} or a perceptual \cite{johnson2016perceptual, poster2021visible} loss into our model might lead to perceptual improvement.}
\begin{comment}
\noindent{\textbf{{Generalization Tradeoff.}}{
{Figure \ref{fig:cold_ablation} and \ref{fig:warm_ablation} show that in contrast to a fully supervised model, model regularization leads to artifact-free generalization on the FAKE dataset. Figure \ref{fig:tufts_side} shows similar findings for non-frontal real-world images. Visually, both models achieve good results on comparable lab-condition images from the TUFTS dataset though, see Figure \ref{fig:tufts}. Nevertheless, 
%Figure  \ref{tab:th_ablation_full} highlights that 
despite good generalization capabilities our final models do not necessarily perform better than supervised models for the thermalization of SEJONG images.  However, this finding
is unsurprising as this is a common consequence of model regularization. Model regularization leads to a bias-variance tradeoff.}
}}
\end{comment}

\noindent{\textbf{Dense Landmarks.} In this work, we relied on the ground truth 70-point landmarks of the FAKE dataset \cite{wood2021fake} and dense Mediapipe \cite{lugaresi2019mediapipe} annotations. Training with the original 320- and 702-point landmarks could further boost accuracy and lead to an even denser landmarker. However, these landmarks are not publicly available. Furthermore, we only evaluate a mobilenet backbone {for landmark detection without face tracking}. {Better} performance for difficult poses and face variability could be achieved with denser models \cite{wood20223d} {and spatial normalization of face positions during training}.

\noindent{\textbf{CHARLOTTE.} The CHARLOTTE \cite{ashrafi2022charlotte} dataset is among the largest datasets with thermal recordings of faces that contain different levels of image quality as well as has a high variability in poses such as side profile pictures and tilting which makes the dataset ideal as a benchmark. %\textcolor{violet}
{However, the {2D} annotation uses {a convention} where side profile images have a different number of landmarks than frontal faces. Furthermore, landmarks of low-resolution images are quantized and there are examples of shifted ground truth annotations (see Figure \ref{fig:sup_landmarks}). } 

\begin{figure*}
\includegraphics[width=\linewidth]{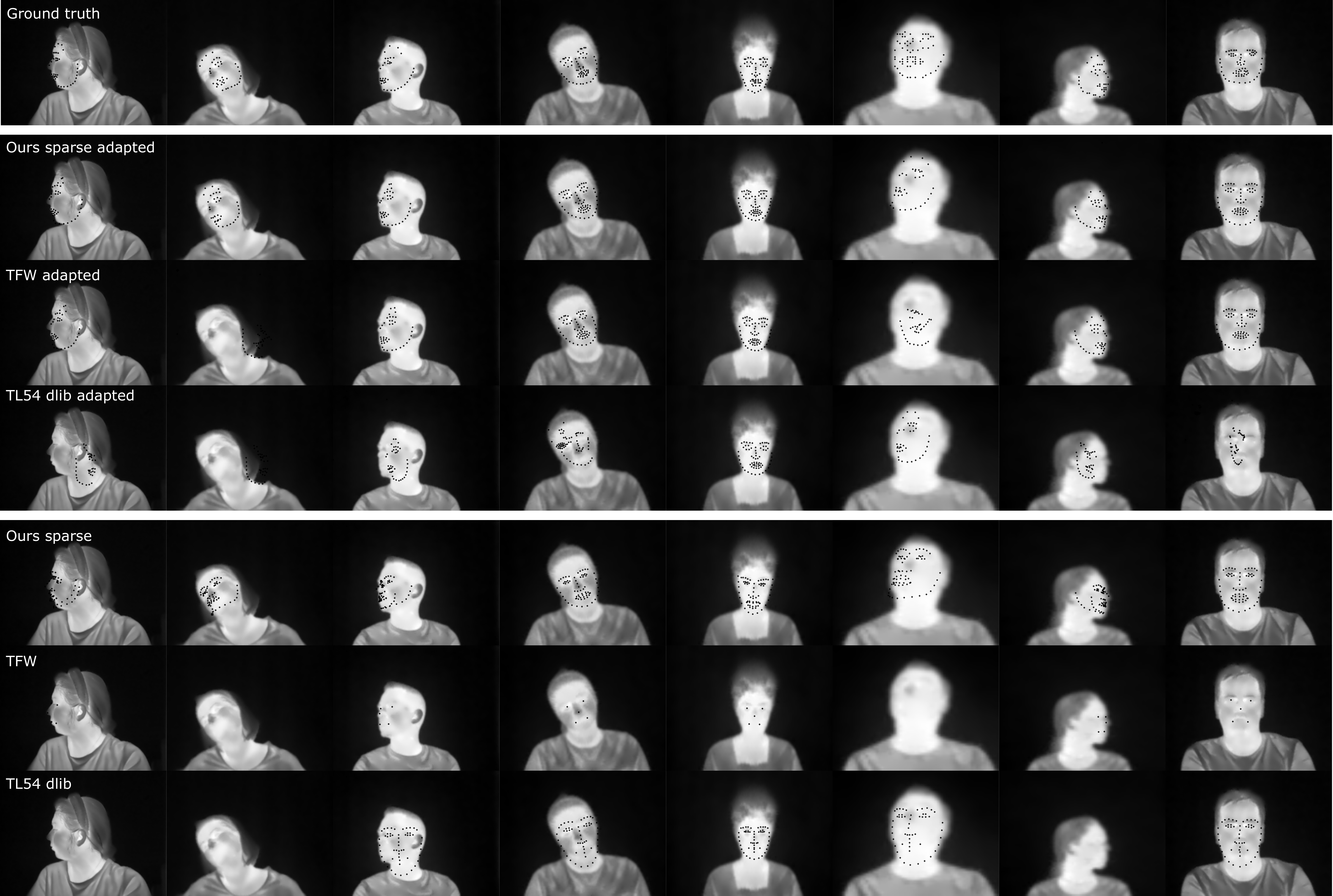}
\caption{Label adaptation examples. For side profile views in the CHARLOTTE dataset only landmarks for visible parts of the face exist and additional positions on the forehead are marked. Label adaptation translates the three different landmark conventions we use for evaluation (bottom) into the CHARLOTTE convention (top).}
\label{fig:sup_adapt}
\end{figure*}

\noindent{\textbf{Label Adaptation.} 
\label{sec:label_adapt}We retrain the label adaptation networks for each tested landmarker on its original predictions. Hence, for a given method, a high failure rate on CHARLOTTE means that fewer training images are available for that method. Also, {poor} predictions on some of the images such as profile images without failure produce a low quality of adapted landmarks (see Figure \ref{fig:sup_adapt}, TL54 dlib, column 3). It is important to note that the label adaptation does not use visual information from the benchmark dataset to translate landmarks. However, the same individuals were present in both the test and the training {datasets} which might allow {networks} to learn {facial statistics} of individuals. This might give an advantage to landmarkers with only a few landmarks, e.g., TFW \cite{kuzdeuov2022tfw} (see Figure \ref{fig:sup_adapt}, TFW). The {good} 68-point landmark performance of TFW on the CHARLOTTE {dataset} does not necessarily mean that this generalizes and the advantage of increasing the number of predicted landmarks has been demonstrated for RGB images \cite{wood20223d}.

\begin{figure*}%[H]
    \centering
    \includegraphics[width=\linewidth]{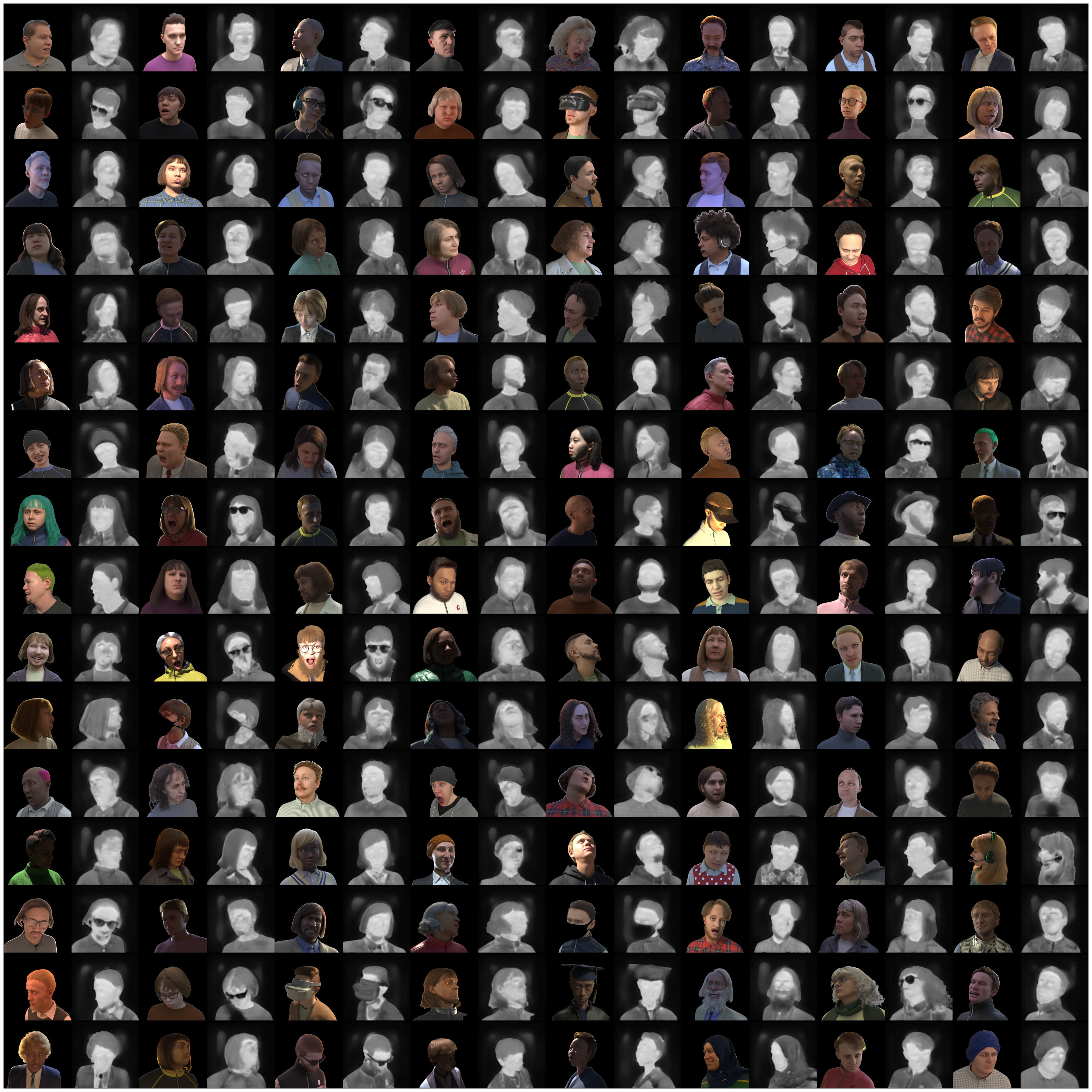}
    \caption{The first 128 FAKE (with removed background) and T-FAKE images with a random choice between the `cold' and the `warm' variant.}
    \label{fig:large}
\end{figure*}

\begin{figure*}%[H]
    \centering
    \includegraphics[width=0.6\linewidth]{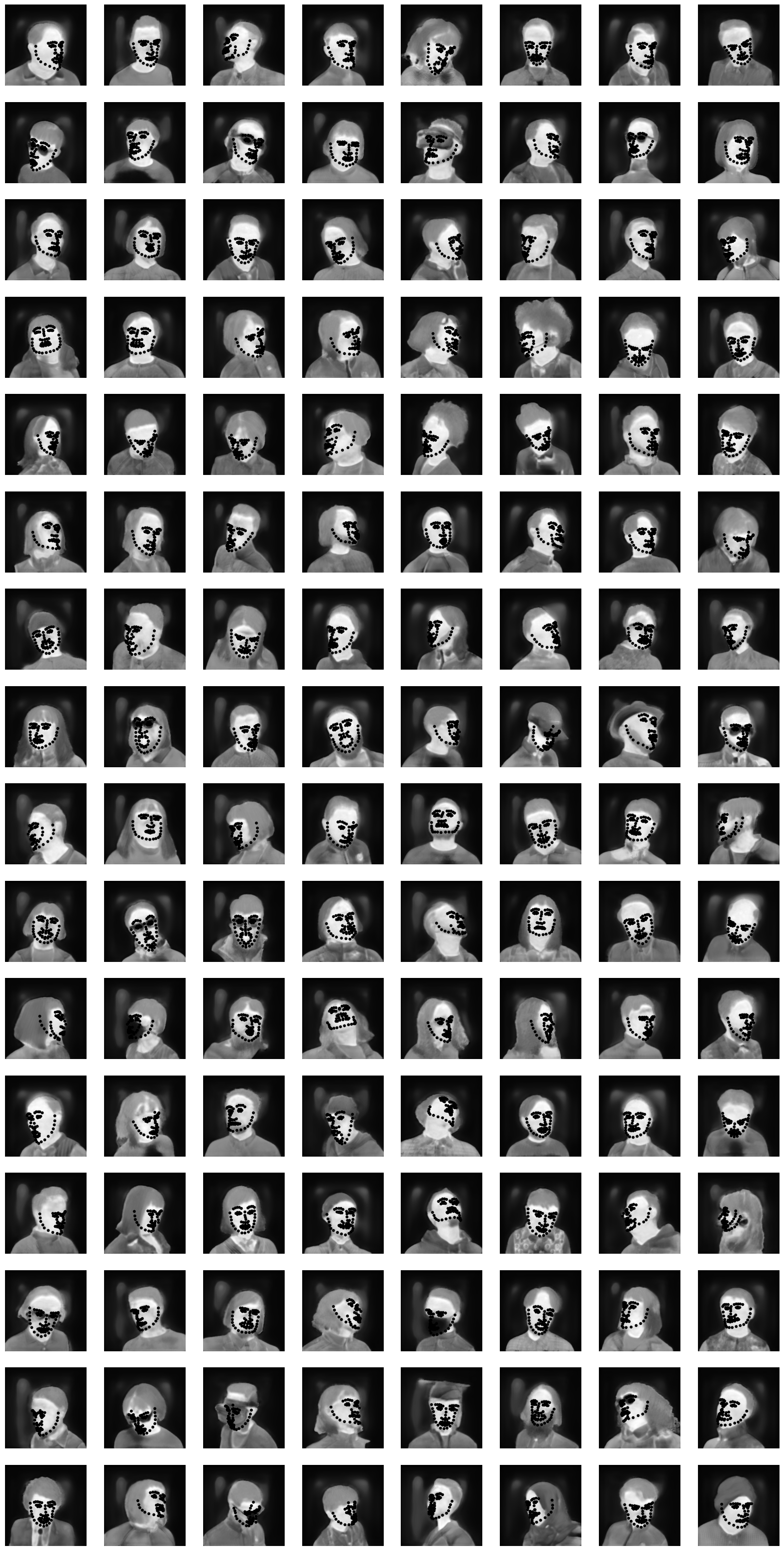}
    \caption{The sparse landmarks for the T-FAKE images in Fig. \ref{fig:large}}
    \label{fig:large_sparselandmarks}
\end{figure*}

\begin{figure*}
    \centering
    \includegraphics[width=0.85\linewidth]{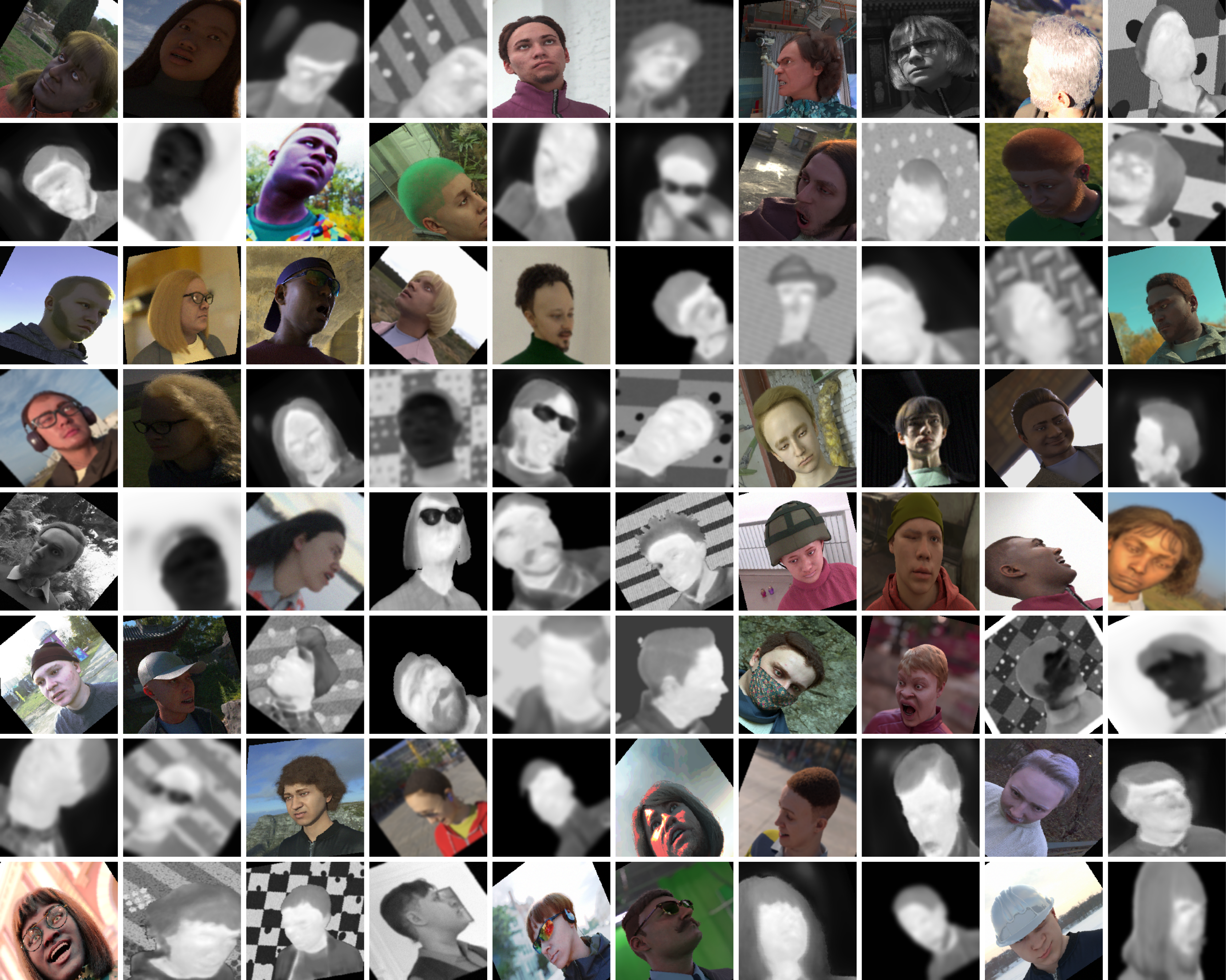}
    \caption{{Example images from FAKE and T-FAKE with the augmentations applied during training.}}
    \label{fig:augmentations}
\end{figure*}

\end{document}